\pdfoutput=1
\documentclass{article}

\usepackage{amssymb}
\usepackage{graphicx}
\usepackage{booktabs}
\usepackage{amsmath}
\usepackage{float}
\usepackage{placeins}
\usepackage{pgfplots}
\usepackage{bbm}        %
\pgfplotsset{compat=1.18}
\usepackage[super]{nth}
\usepackage{subcaption}

\newcommand{\deepmonocularmodel}{SMTS} %

\usepackage[final]{corl_2026} %

\title{Mechanistic Interpretability for End-to-End Self-Driving}

\author{
\normalfont
\textbf{
Manav Gagvani$^{1}$,
Benjamin Namikas$^{2}$,
Sivamurugan Velmurugan$^{2}$,
Nysa Kumar$^{2}$
} \\
\textbf{
Shrey Sharma$^{2}$,
Thomas Peterson$^{2}$,
Pratyush Mathur$^{2}$,
Hyunseo Chang$^{2}$,
Rohan Paleja$^{2}$
} \\[2mm]
$^{1}$Department of Electrical \& Computer Engineering, Purdue University \\
$^{2}$Department of Computer Science, Purdue University \\
{\small\texttt{\{mgagvani, bnamikas, svelmuru, kumar753,}} \\
{\small\texttt{shar1159, peter922, mathur91, chang899, rpaleja\}@purdue.edu}}
}

\hypersetup{
  pdftitle={Mechanistic Interpretability for End-to-End Self-Driving},
  pdfauthor={Manav Gagvani, Benjamin Namikas, Sivamurugan Velmurugan, Nysa Kumar, Shrey Sharma, Thomas Peterson, Pratyush Mathur, Hyunseo Chang, Rohan Paleja}
}

\begin{document}
\maketitle

\begin{abstract}
    End-to-end (E2E) autonomous driving models achieve strong planning performance but remain largely opaque, limiting observability and trust. We apply mechanistic interpretability to E2E driving by training Sparse Autoencoders (SAEs) on the internal activations of a simple, single-camera end-to-end path planning model. Unsupervised clustering of SAE latents reveals interpretable structure aligned with route intent, vehicle speed, and braking behavior, recovering route intent with 98.3\% accuracy without supervision. Through paired counterfactual image edits, we further show that specific SAE features are visually grounded and systematically respond to scene changes such as traffic light state and stop sign presence. Steering individual SAE latents produces predictable, monotonic changes in planner outputs including lateral displacement, curvature, and braking. Finally, we demonstrate a practical application of these representations: we build a lightweight adapter trained on SAE latents that reduces trajectory selection error by 3.44\%, outperforming an equivalent adapter that omits SAE latents, confirming that the sparse representation captures useful information for interpretability. Code and visualizations are available at
\url{https://mgagvani.github.io/mechanistic-interpretability-for-e2e-driving/}.
\end{abstract}


\section{Introduction}

Road safety remains one of the most urgent motivations for autonomous driving research. Each year, road traffic kills approximately 1.2 million people worldwide, making autonomous vehicles both a research and public health problem \citep{who2023globalroadsafety, slotkin2025selfdrivingcars}. Companies such as Wayve are deploying learned end-to-end (E2E) autonomy models on public streets \citep{wayve2023lingo1, marcu2024lingoqavisualquestionanswering}, but as E2E autonomous driving systems transition from simulation to deployment in the real world, post-deployment safety through observability and understanding of model behavior becomes increasingly important.

Over the past decade, the field of autonomous driving (AD) has shifted from modular pipelines with handcrafted perception, planning, and control modules, to end-to-end models that directly learn from data to map raw sensor data and ego state into driving decisions \citep{challengesandfrontiers, paleja2023learninginterpretablehighperformingpolicies}. Despite advancements in the application of Vision-Language-Action (VLA) models for autonomous driving \citep{renz2025simlingovisiononlyclosedloopautonomous, sima2025drivelmdrivinggraphvisual, ding2024hintadholisticallyalignedinterpretability}, the black-box nature of E2E models diminishes trust and restricts practical applications due to limited safety bounds and understanding of model behavior in diverse or out-of-distribution scenarios. To address this shortcoming, some E2E methods produce intermediate outputs such as bird's-eye-view occupancy or motion prediction \citep{hu2023planningorientedautonomousdriving, paradrive}, providing clues as to a model's understanding of a scene but little insight within the exact inner workings of the network.

To better understand E2E models, we leverage techniques from mechanistic interpretability \citep{gao2024scalingevaluatingsparseautoencoders, lan2025quantifyingfeaturespaceuniversality}. Specifically, we apply Sparse Autoencoders (SAEs), an unsupervised learning technique that disentangles polysemantic representations by projecting them into a sparse latent space. This technique has been used to find highly interpretable features in LLMs \citep{cunningham2023sparseautoencodershighlyinterpretable} and visuomotor VLA policies \citep{swann2026sparseautoencodersrevealinterpretable, jin2026eventgroundedsparseautoencodersvisionlanguageaction}. To our knowledge, we are the first to apply these techniques to E2E driving policies.

Our contributions are as follows:

\begin{enumerate}
    \item An elegant, high-performance end-to-end planning module using trajectory scoring, without any intermediate labeled representations or large language modeling backbones, that achieves competitive open-loop planning performance on the Waymo Open Dataset's E2E driving benchmark \citep{xu2025wode2ewaymoopendataset}.
    \item We show that trained SAE features organize into interpretable clusters aligned with route intent and kinematic behavior, and that these features respond systematically to counterfactual visual changes in driving scenes.
    \item We show that SAE latents are useful representations for model steering and improvement. Latent steering produces structured changes in planner behavior consistent with clustering and image edit analyses, and using SAE latents as inputs to an adapter subnetwork improves model performance over the original dense query activation.
\end{enumerate}

\section{Related Work}

Our work combines three areas of research: end-to-end autonomous driving, interpretability for driving policies, and mechanistic interpretability through sparse feature learning. %

\textbf{End-to-End Autonomous Driving --}
Since the seminal paper, ``End to End Learning for Self-Driving Cars'' \citep{bojarski2016endendlearningselfdriving}, which trained a small CNN to map front-facing camera images to steering commands, there has been an explosion of model architectures for the task of end-to-end autonomous driving. Two major categories of trajectory generation methods are direct trajectory regression and trajectory scoring \citep{challengesandfrontiers, jiang2026vadv2endtoendvectorizedautonomous, li2025ztrszeroimitationendtoendautonomous}. One key baseline model is TransFuser \citep{chitta2022transfuserimitationtransformerbasedsensor, jaeger2023hiddenbiasesendtoenddriving, nguyen2026leadminimizinglearnerexpertasymmetry} which fuses multiple cameras and LiDAR point cloud inputs by using a cross-attention feature pyramid architecture. Regressing a single trajectory limits the multimodality of the planner's output, leading to the development of trajectory scoring methods \citep{xing2025goalflowgoaldrivenflowmatching}. This category can be further split into networks which solely use a pre-computed vocabulary of actions \citep{li2024hydramdpendtoendmultimodalplanning, li2025hydramdpadvancingendtoenddriving, yao2025drivesuprimprecisetrajectoryselection}, and methods which dynamically generate trajectory proposals to be selected from \citep{kirby2026drivingregisters, li2025generalizedtrajectoryscoringendtoend, guo2025ipaditerativeproposalcentricendtoend}. Our model falls under the class of trajectory scoring approaches, specifically methods that dynamically generate trajectory proposals and then select among them using a learned scoring head. Our model is unique in that it does not incorporate intermediate labeled representations such as HD maps or bird's-eye-view occupancy, and only uses a single camera as visual input. 

\textbf{Interpretability in End-to-End Autonomous Driving --}
Existing E2E models approach interpretability through auxiliary output heads or textual reasoning, but do not identify internal features used by the planner to generate or select trajectories \citep{nvidia2026alpamayor1bridgingreasoningaction, paradrive}. Existing work exposes intermediate outputs such as motion prediction and predicted scene occupancy \citep{hu2023planningorientedautonomousdriving, paradrive}, helping understand what the model's internal representations can extract from given inputs. More recently, vision-language models and VLA models have been used to improve the transparency and interpretability of E2E autonomous driving models \citep{renz2025simlingovisiononlyclosedloopautonomous}. For instance, DriveLM and the Alpamayo-R1 VLA use large backbones to perform visual question answering \citep{sima2025drivelmdrivinggraphvisual, nvidia2026alpamayor1bridgingreasoningaction}. Hint-AD inserts a pretrained vision-language model into existing E2E architectures, producing reasoning and 3D dense captioning which are aligned to the model's internal representation \citep{ding2024hintadholisticallyalignedinterpretability}. These existing methods improve interpretability by exposing additional model outputs, such as occupancy, motion prediction, captions, or language-based reasoning. In contrast, our work focuses on the planner’s internal representations: we study whether the internal features used to produce trajectories can be identified and linked to driving behavior.

\textbf{Mechanistic Interpretability and Sparse Autoencoders --}
Mechanistic interpretability studies internal model representations rather than input-output behavior alone. A key obstacle is \textit{polysemanticity}: individual neurons often respond to multiple unrelated concepts, likely because models store features in superposition \cite{elhage2021mathematical, elhage2022toymodelssuperposition}. Sparse autoencoders (SAEs) address this by learning a dictionary that reconstructs hidden activations with sparse coefficients, yielding features that are more monosemantic and interpretable than raw neurons \cite{cunningham2023sparseautoencodershighlyinterpretable}. Recent work replaces $L_1$ sparsity penalties with TopK activations, which directly control the number of active latents, and adds an AuxK loss to reduce dead latents~\citep{gao2024scalingevaluatingsparseautoencoders}.
 Critically, \citet{marks2025sparsefeaturecircuitsdiscovering} demonstrate that sparse feature circuits can be identified and edited to steer model behavior. Most relevant to the E2E autonomous driving setting, \citet{swann2026sparseautoencodersrevealinterpretable} extend this framework, training SAEs on VLA robot manipulation policies and demonstrating that individual features can causally steer trajectory behavior. These works show that SAEs can expose interpretable and causally meaningful features in language models and VLA policies. However, it remains unclear whether the same approach can reveal useful internal structure in E2E autonomous driving models, which can lack language modeling backbones. Our work addresses this by applying SAEs to the internal activations of an E2E driving model and evaluating whether the resulting features are aligned with kinematic and visual features, and if those features can alter model behavior.

\section{Method}

\subsection{End-to-End Driving Model}

\begin{figure}[htpb]
    \centering
    \includegraphics[width=0.9\linewidth]{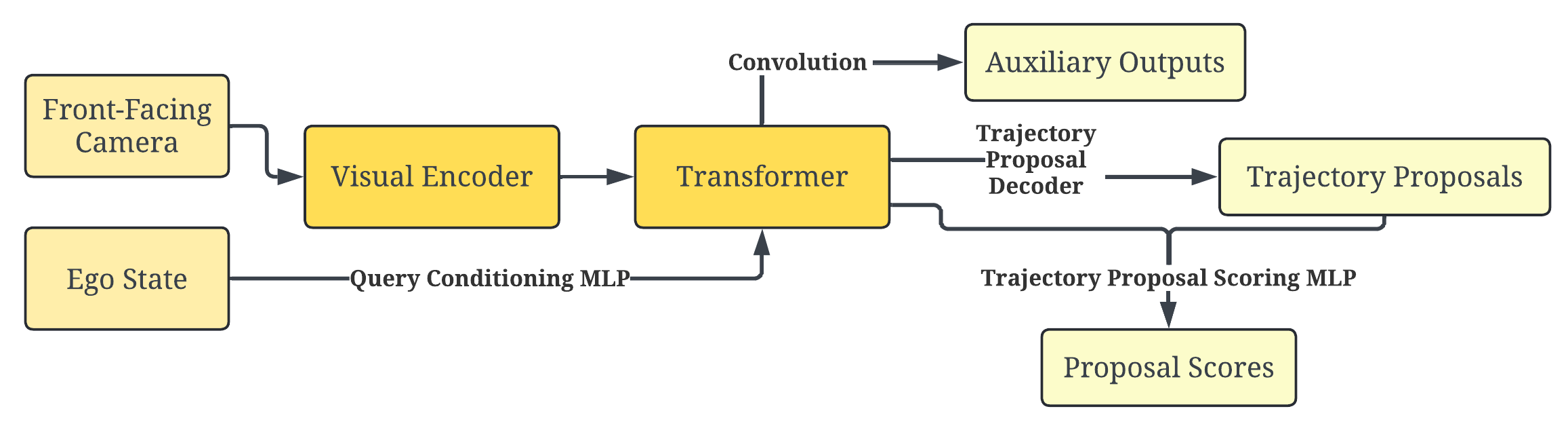}
    \caption{\textbf{Model Architecture.} Our Simple Monocular Trajectory Scorer (\deepmonocularmodel) uses only the front camera, past ego state, and route intent as input. Past state and intent initialize a query token, which attends to adapted visual tokens through four transformer blocks. The final query predicts candidate control sequences, rolls them out into future trajectories using a kinematic motion model, and scores each trajectory with a learned proposal scorer. A supervised depth head is used as an auxiliary training output. }
    \label{fig:deepmonocular_arch}
\end{figure}

\FloatBarrier

We develop an elegant, high-performance E2E architecture, the Simple Monocular Trajectory Scorer (\deepmonocularmodel) shown in Figure \ref{fig:deepmonocular_arch}. Many existing methods either use large VLM/VLA modules or incorporate diverse sensor inputs such as LiDAR or multiple cameras. To isolate the internal planning representation, we use a monocular E2E planning model which consists of four stages: visual encoding, query conditioning, trajectory proposal decoding, and trajectory proposal scoring.

\textbf{Visual encoding --} Building on transformer backbones used in \citet{kirby2026drivingregisters}, the model first extracts patch-level features from the front-facing camera image using a vision transformer (ViT) backbone. We freeze the backbone during training and learn a lightweight convolutional visual adapter on top of its feature map, which is then flattened into $N_t$ visual tokens each of dimension $d$. An auxiliary dense depth head encourages the adapter to preserve geometric information useful for planning.

\textbf{Query conditioning --} \citet{li2024egostatusneedopenloop} and \citet{zhai2023ADMLP} showed that past ego status alone can achieve high performance on open-loop autonomous driving benchmarks, creating a risk that the model learns shortcuts that bypass visual scene understanding. To make this information available while keeping it localized, we inject past ego motion and route intent only through the initial query. Concretely, route intent is converted to a one-hot vector and concatenated with the past ego state to produce a vector $v$. A learned linear projection maps $v$ to an initial query $Q \in \mathbb{R}^d$. This query is the only token updated by the transformer decoder, so the decoder must fuse $v$ with visual tokens through attention rather than directly emitting trajectories conditioned on the ego state.

\textbf{Trajectory proposal decoding --} Before decoding proposed trajectories, we add a learned positional encoding to each token. Next, a transformer decoder, which is a stack of 4 transformer blocks, produces the final query $Q'$. An MLP transforms $Q'$ to a set of $N_p$ trajectory proposals, each of which is a sequence of $T$ poses predicted from the current timestep $t$ up to a future horizon at $t+T$. Each timestep consists of $(\dot{v}, \omega) \in \mathbb{R}^2$, which are rolled out into $(x,y) \in \mathbb{R}^2$ future poses via a kinematic model in the ego agent's local reference frame at timestep $t$.

\textbf{Trajectory proposal scoring --} The scoring head assigns a scalar cost to each decoded proposal rather than producing a single trajectory directly. Each proposed $(x,y)$ trajectory is first flattened and embedded with an MLP. The resulting trajectory embedding is concatenated with the detached final planner query $\text{Detach}(Q')$. A second MLP maps this concatenated representation to one scalar score per proposal. At inference, the model selects the proposal with the lowest score, yielding the predicted best trajectory.

\subsection{Sparse Autoencoder with AuxK Loss}

To analyze the internal planning representation of \deepmonocularmodel, we train sparse autoencoders using the TopK formulation and AuxK auxiliary loss, following prior work on VLA interpretability~\citep{gao2024scalingevaluatingsparseautoencoders, swann2026sparseautoencodersrevealinterpretable}. We apply the SAE to the query activations of our compact end-to-end planning model. Since the transformer query concentrates route intent, past ego motion, and visual evidence before proposal generation and scoring, it provides a compact target for mechanistic analysis.

For each transformer block $\ell \in \{0,1,2,3\}$, we extract the query activation after the block $x_\ell \in \mathbb{R}^d$ and train a separate SAE for that block. Each SAE first normalizes activations by subtracting the training-set geometric median, removing the per-sample feature mean, and applying $L_2$ normalization. The normalized activation $\tilde{x}$ is encoded with a TopK sparse nonnegative code, producing the latent $z$ in Equation \eqref{eq:sae_one}.
\begin{equation}
   z = \mathrm{ReLU}\left(\mathrm{TopK}(W_{\mathrm{enc}}\tilde{x}, k)\right).
\label{eq:sae_one}
\end{equation}
The decoder reconstructs the normalized activation $\hat{x}$ as $W_{\mathrm{dec}}z$. Finally, the training objective is $\mathcal{L}_{\mathrm{rec}} + \alpha_{\mathrm{aux}} 
\mathcal{L}_{\mathrm{aux}}$, where $\mathcal{L}_{\mathrm{rec}}$ is normalized 
reconstruction error, $\mathcal{L}_{\mathrm{aux}}$ is the AuxK residual loss over inactive or dead latents, and $\alpha_{\mathrm{aux}}$ is a training hyperparameter. The resulting sparse latents are used as the feature space for clustering, counterfactual visual edit analysis, latent steering, and scorer correction.

\subsection{SAE Analysis and Model Steering} 
\label{sec:methods_33}

After training SAEs on SMTS query activations, we analyze the learned latent space at four levels: whether features organize into coherent groups, whether latents are directly affected by visual input, whether individual features respond to controlled visual changes, and whether intervening on those features causally changes planner behavior. We then use the same latent representation in a lightweight scoring adapter to test whether the sparse features contain information useful for improving trajectory selection.

We first analyze co-activation structure among SAE latents. We perform this analysis on the SAE latent $z_3$ from the final transformer block because this representation is the planner query immediately before trajectory proposal decoding and scoring.

Using the SAE activations from the final block, $z_3$, we first filter individual features by activation rate, retaining those with activation frequencies in $[f_1, f_2]$. This filters out latents which are exceedingly rare and latents which activate on most scenes.
For the retained features, we compute the pairwise Pearson correlation of binarized SAE activations $\mathbf{1}[z_{ij} > 0]$ over the full validation set. We then construct an undirected graph whose nodes are retained SAE features, with an edge between features $a$ and $b$ when their correlation $\rho_{ab}$ exceeds a threshold $\tau$. Connected components of this graph with at least $m$ features define feature clusters. To interpret a cluster $C$, we rank scenes in the validation set by activation score 
$\sum_{j \in C} z_{3,j}(i)$ and inspect the highest-scoring scenes.

These are scenes in which many features in the cluster fire strongly, making them natural candidates for interpreting the shared factor represented by the cluster. To characterize each cluster, we summarize scene, trajectory, scoring, and error statistics over the validation set. We compute z-scores using these statistics, indicating which driving conditions and planner behaviors are overrepresented among scenes selected by the cluster. Using the 50 highest-scoring scenes, we define each cluster's semantic label as the combination of the most common route intent and the sign of the mean speed z-score in that representative subset.

After identifying co-activated feature groups, we next test whether SAE features are grounded in the visual scene rather than only reflecting ego motion or route intent. Since the query is initialized from past ego motion and route intent, clustering structure could be affected by metadata inputs. To disambiguate this, we quantify how latent variation depends on the input image, intent, and ego history by systematically recombining these inputs. For a set of validation samples, we inference the model over the sample's image and six different route intent and ego history combinations. Using the resultant set of latent vectors, we perform an ANOVA variance decomposition, separating the variation in latents into the contributions and interactions between the three model inputs. 

To identify SAE features which correspond with items relevant to driving, we construct paired counterfactual image edits by holding all inputs constant while only changing the input image, producing matched pairs where changes in the SAE activations are solely due to the visual change. For each scene, a YOLO object detector provides context to a VLM prompt generator, which produces 
scene-consistent edit prompts (e.g., only adding pedestrians where a crosswalk is 
visible). An image editing model applies each prompt, forming matched pairs 
$(I, p, r)$ and $(I', p, r)$ with identical past state $p$ and intent $r$, 
yielding query tokens $(h, h')$ and latents $(z, z')$.

We quantify each edit via token-level shift $\lVert h' - h \rVert_2$, sparse latent 
shift $\lVert z' - z \rVert_2$, and signed per-feature delta $z' - z$. We then 
validate feature monosemanticity by directly steering individual latent features 
via $z_i' = \max(0, z_i + \alpha s_i)$, where $s_i$ is a per-feature normalization 
factor, and measuring downstream planner behavior.

For each feature, we select scenes in which it is most active and sweep $\alpha$ over a fixed set of values $\mathcal{A}$, producing a modified latent $\tilde{z}$ and updated final query $\tilde{Q}$, from which new proposals and corresponding scores are computed. At each intervention level $\alpha_j \in \mathcal{A}$, we measure how the intervention changes both the selected trajectory and the proposal distribution. Specifically, we track lateral displacement, curvature, braking and acceleration magnitude, score margin, and proposal spread. These metrics, along with feature quality, are defined in Appendix~\ref{sec:supp_latent_steering}.

Finally, we ask whether the SAE latent space is not only interpretable, but also useful for improving the planner’s trajectory selection. This latent representation can be used to improve the performance of the scoring submodule after training without requiring supervised fine-tuning of the full model. We propose an auxiliary adapter module to predict the original scoring head error $\epsilon_{\mathrm{scorer}}$ using SAE latent activations. Because model error largely stems from misclassifying the best generated trajectory, we target the scoring submodule directly. The adapter predicts each proposal’s scorer error using an $L_1$ loss. At inference time, it corrects predicted scores by adding predicted error.

\section{Results}

We present results demonstrating that 1) \deepmonocularmodel\ is comparable to 
existing E2E planners and 2) that extracted SAE features are both \textbf{interpretable} 
and \textbf{steerable}: we cluster SAE features to reveal interpretable structure, 
use counterfactual image edits to show features are visually grounded, demonstrate 
that individual features can predictably alter planner output, and show that SAE 
features improve downstream performance via a learned adapter.

\subsection{Open-Loop Planning}
All experiments use a single front-facing camera as input to a frozen pretrained 
Perception Encoder (ViT-S) \citep{bolya2025PerceptionEncoder}, with decoder and 
SAE hidden dimension 384, trained on 250,000 WOD-E2E 
\citep{xu2025wode2ewaymoopendataset} samples for 10 epochs with a learning rate of $1\times10^{-4}$.
WOD-E2E mines rare long-tail scenarios (construction, intersections, multi-lane 
maneuvers) absent from existing benchmarks \citep{xu2025wode2ewaymoopendataset, nuscenes}. Our \deepmonocularmodel\ model achieves an ADE of 1.40~m on the validation split, outperforming Latent TransFuser (1.50~m), GTRS-Dense 
(1.48~m), and DrivoR (1.49~m) \citep{chitta2022transfuserimitationtransformerbasedsensor, 
li2025generalizedtrajectoryscoringendtoend, kirby2026drivingregisters}. On the private WOD-E2E test evaluator, SMTS also achieves the highest average Rater Feedback Score (7.765 vs. 7.688 for GTRS-Dense and 7.645 for DrivoR), leading on 6 of 11 scene categories. We further show that \deepmonocularmodel \ scales more efficiently with limited data compared to competing methods, as well as full WOD-E2E results, in Appendix \ref{sec:app_open_loop}.

\subsection{Unsupervised Feature Clustering}
\label{sec:feature_clustering}

\begin{figure}[htpb]
    \centering
    \includegraphics[width=1\linewidth]{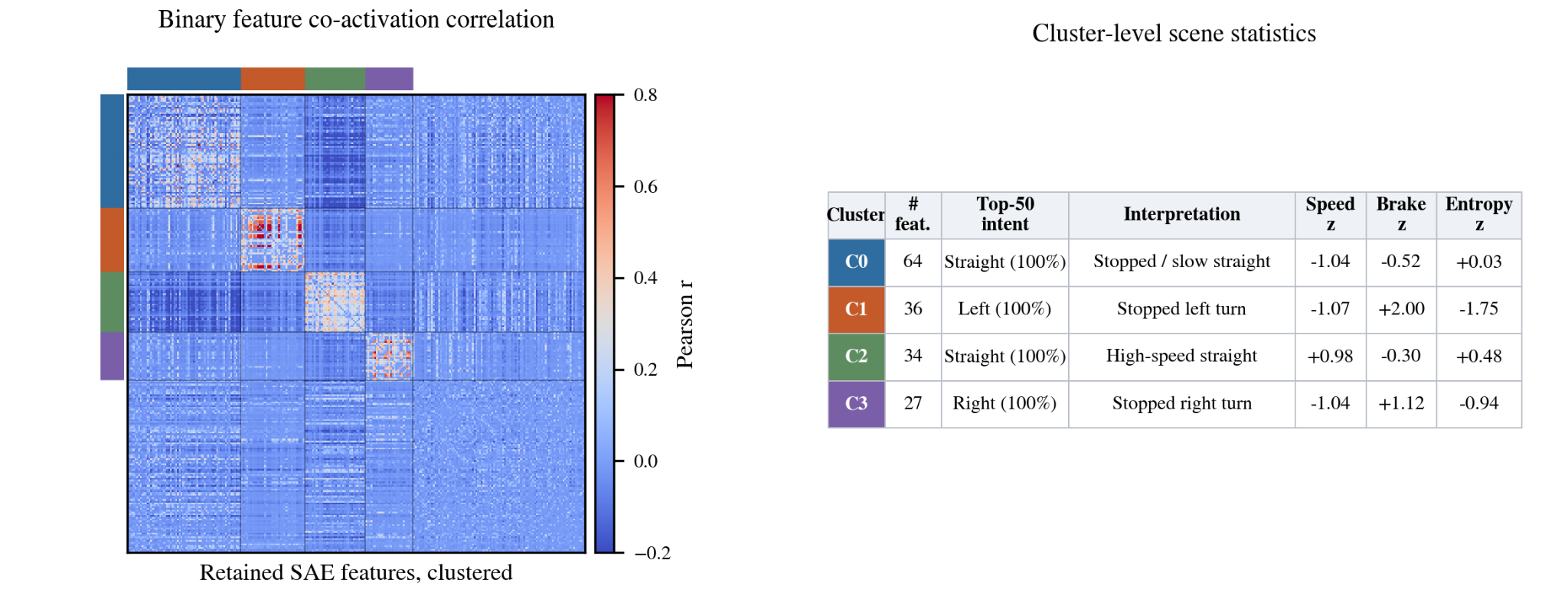}
    \caption{
\textbf{Unsupervised SAE feature clusters.}
\textit{Left}: Pairwise Pearson correlation matrix of binarized SAE latent activations. Features are ordered by connected components, revealing four major clusters. \textit{Right}: Cluster-level summaries for the connected components obtained at $\tau=0.35$, including cluster size, dominant route intent, and standardized driving/model statistics for the scenes that most strongly activate each cluster. The clusters separate left- and right-turn behavior from driving straight, while the two straight clusters further separate low-speed and high-speed scenes.
}
    \label{fig:feature_correlation}
\end{figure}

Next, we follow the clustering procedure in Section~\ref{sec:methods_33} to test whether the learned SAE latents exhibit organized structure. We analyze the SAE latents from the final transformer block, $z_3$, over the full WOD-E2E validation set. Using pairwise Pearson correlations with $\tau=0.35$, we construct a co-activation graph by retaining connected components of at least three features, yielding four clusters of 64, 36, 34, and 27 latents (Figure~\ref{fig:feature_correlation}). The same four groups are recovered by greedy modularity maximization
\citep{clauset2004finding}, a threshold-free alternative to connected components, with 94.3\% of features receiving the same label (Appendix~\ref{sec:supp-cluster-threshold}).%

To quantify alignment with semantic driving structure, we assign each scene to the cluster 
with the highest mean activation $|C_k|^{-1}\sum_{j \in C_k} z_{3,j}(i)$, and 
map clusters to intents by majority vote. This assigns $C_0$, $C_2$ to driving straight, 
$C_1$ to left turns, and $C_3$ to right turns, recovering route intent with 98.3\% accuracy 
overall (per class accuracies are 99.8\% for straight, 92.9\% for left turns, and 86.4\% for right turns).

Among the top 50 scenes ranked by cluster activation, $C_0$ corresponds to stopped or slow straight-driving scenes, with mean speed 1.04 standard deviations below the validation mean, while $C_2$ corresponds to high-speed straight-driving scenes, with mean speed 0.98 standard deviations above (Appendix~\ref{sec:supp-cluster-threshold}--\ref{sec:supp-cluster-scenes}). Thus, the co-activation structure recovers both the explicit route intent variable and an additional continuous driving factor, speed, without using either signal to construct the clusters. This unsupervised alignment suggests that the SAE has learned an internally organized representation of planning-relevant scene structure.

\subsection{Input Conditioning Analysis}
\label{sec:variance_results}
Since the query is initialized from route intent and past ego motion, we test whether the final transformer block's latent representation remains dependent on visual input when these inputs are controlled. We evaluate 9,000 validation images under a combination of three route intents and two fixed ego histories, and apply an ANOVA decomposition to the resulting latent activations. Table~\ref{tab:variance-decomposition} shows that the image accounts for 14.3\% of latent variance directly, and a further 18.5\% through interactions with intent and ego history. The image $\times$ intent and image $\times$ ego history terms ($7.2\%$, $6.9\%$) are significant, since if the latents simply re-encoded these inputs with no contribution from the image, these terms would be zero. The latents therefore combine visual evidence with intent and ego state rather than copying them forward.

\begin{table}[h]
\centering
\caption{Share of SAE latent variance explained by each input.}
\label{tab:variance-decomposition}
\small
\begin{tabular}{@{}lr@{\hspace{1.5em}}lr@{\hspace{1.5em}}lr@{}}
\toprule
Source & Var. (\%) & Source & Var. (\%) & Source & Var. (\%) \\
\midrule
Image       & 14.3\% & Intent $\times$ ego history   & 16.9\% & Image $\times$ intent $\times$ ego history & 4.4\% \\
Intent      & 26.1\% & Image $\times$ intent & 7.2\%  &                                    &       \\
Ego history & 24.0\% & Image $\times$ ego history    & 6.9\%  &                                    &       \\
\bottomrule
\end{tabular}
\end{table}

\subsection{Counterfactual visual edits}
\FloatBarrier
\label{sec:counterfactual_results}

Next, we show that SAE features are visually grounded, rather than merely reflecting simple kinematic features such as route intent. We use the procedure described in Section~\ref{sec:methods_33} to perform this analysis: holding each sample's past ego state and route intent fixed while modifying only the front-facing camera image, and then measuring changes in the output trajectory and SAE latents.

\begin{figure}[h]
    \centering
    \includegraphics[width=1\linewidth]{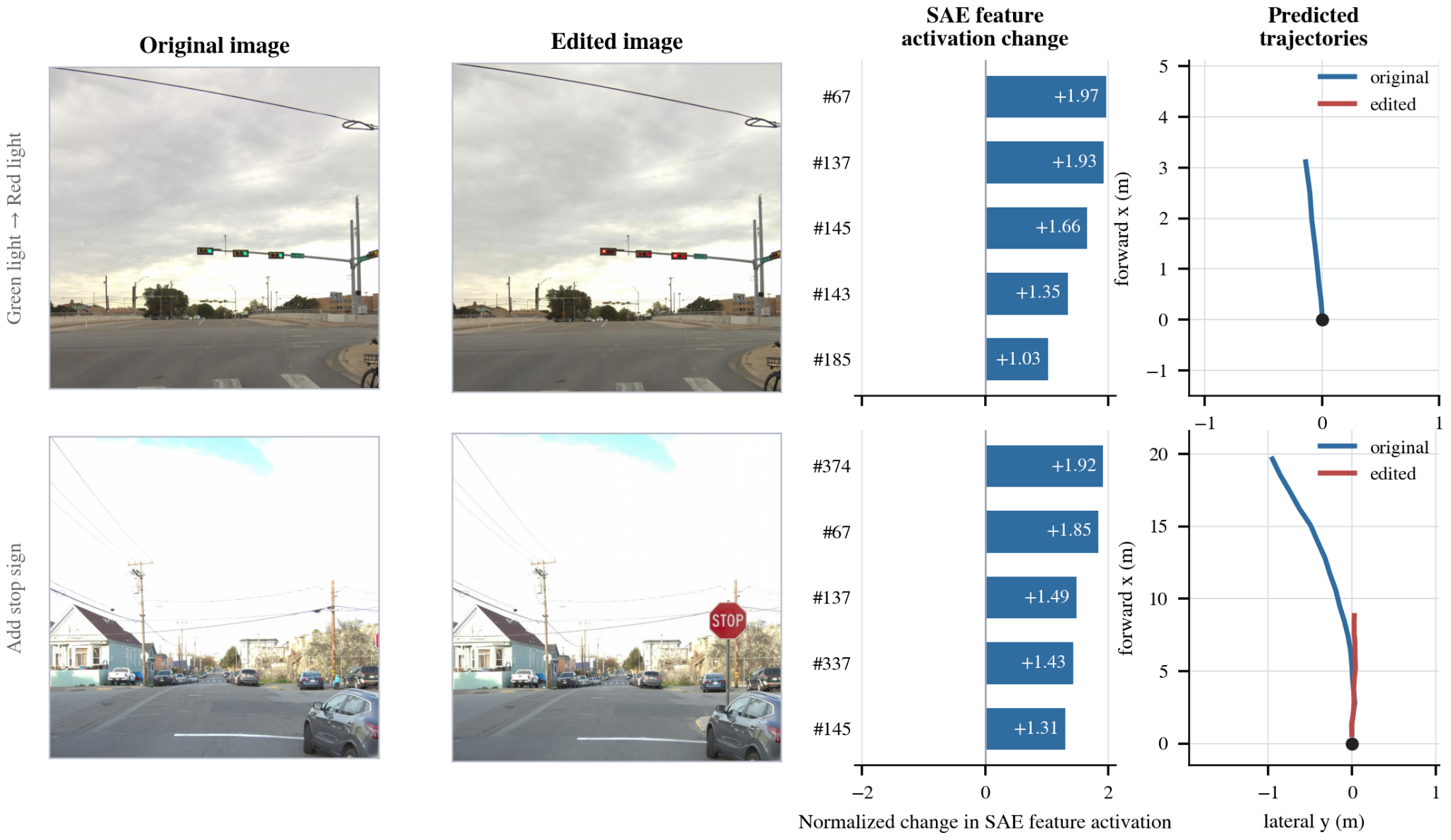}
    \caption{
\textbf{Counterfactual visual edits reveal visually grounded SAE features.}
Each row shows a paired intervention in which the front-facing image is edited while ego history and route intent are held fixed. From left to right, columns show the original image, edited image, largest normalized changes in SAE feature activations, and the selected trajectory before and after the edit. \textit{Top}: changing green traffic lights to red produces large shifts in a small set of SAE features and changes the selected trajectory from forward motion to stopping. \textit{Bottom}: inserting a stop sign reduces the selected trajectory displacement. 
}
    \label{fig:counterfactual}
\end{figure}

Figure \ref{fig:counterfactual} shows two representative edits. Each row shows the original
image, the edited image, the top 5 SAE feature activation changes (normalized by each feature's standard deviation across the validation set), and the resulting selected trajectory overlaid on the original. In the first edit, switching a green traffic light to red produces a shift in specific SAE features, and changes the predicted trajectory from forward motion to a full stop. In the second edit, inserting a stop sign into the scene produces a different shift in SAE feature activations, and also decreases the predicted displacement of the vehicle.

These examples demonstrate that SAE features capture semantically meaningful visual changes that affect trajectory selection, motivating our latent steering analysis in Section~\ref{sec:latent_steering_results}.

\subsection{Latent Steering}
\label{sec:latent_steering_results}

\begin{figure}[htbp]
    \centering
    \includegraphics[width=1\linewidth]{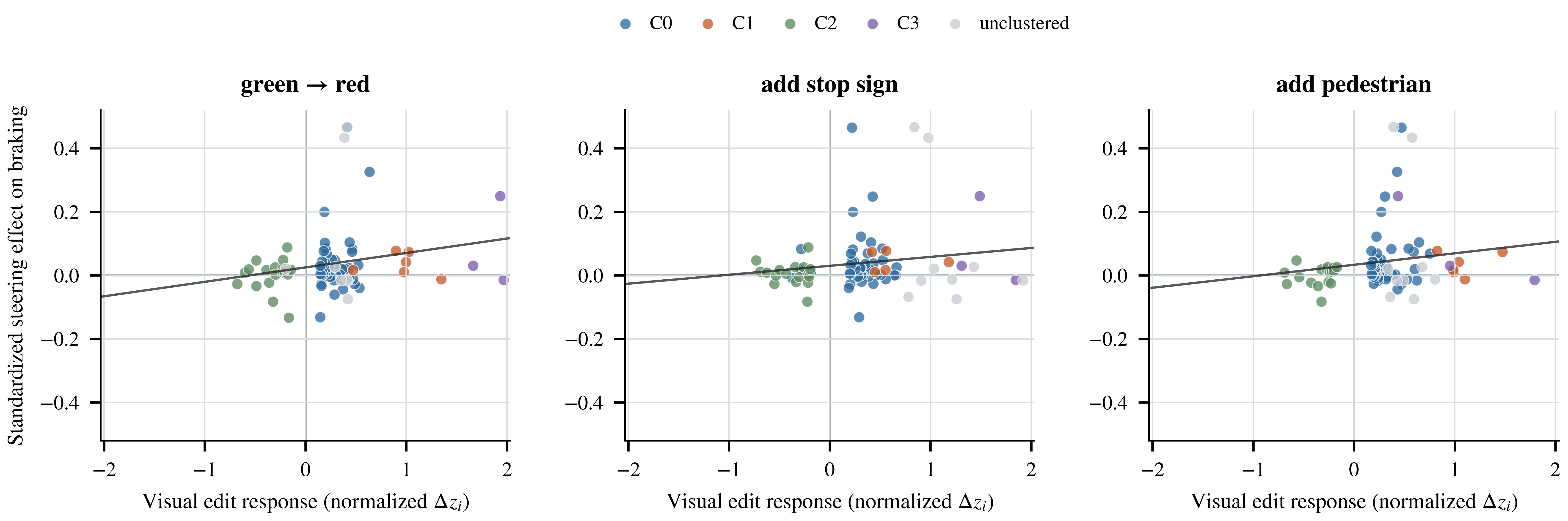}
    \caption{
    \textbf{SAE features have aligned steering effects.}
    Each panel compares counterfactual visual sensitivity with latent steering. Each point is an SAE feature selected among the most responsive features for that edit. The x-axis shows the feature's normalized signed activation change under the visual edit, while the y-axis shows the standardized change in braking magnitude when the same feature is manually increased through latent steering. Across green-to-red traffic light edits, added stop signs, and added pedestrians, relevant features show a consistent positive alignment between visual response and expected braking behavior.
    }
    \label{fig:alignment}
\end{figure}

To connect visual sensitivity with steering behavior, we compare two independently computed quantities for each SAE feature. For an edit direction $e$, we replay the original and edited images through the planner, encode the resulting query activations with the SAE, and define $x_{e,i}$ as the feature's mean signed activation change across successful edits divided by its validation-set standard deviation, shown in Figure \ref{fig:alignment}.

We compare each feature’s visual response to its direct steering effect on braking. For feature $i$, let $y_i$ be the mean standardized change in braking after amplifying that feature and rerunning the planner. Figure~\ref{fig:alignment} plots $y_i$ against the visual edit activation response $x_{e,i}$ for the most relevant features under each stop-inducing edit. Alignment is consistently positive across green-to-red light edits, added stop signs, and added pedestrians: 61\% of interventions match the visual edit sign, with Pearson $r=0.172$. This indicates that visually responsive SAE features also tend to produce the expected trajectory change when steered directly.

\subsection{Lightweight Scorer Adaptation}

\begin{figure}[h]
    \centering
    \includegraphics[width=1\linewidth]{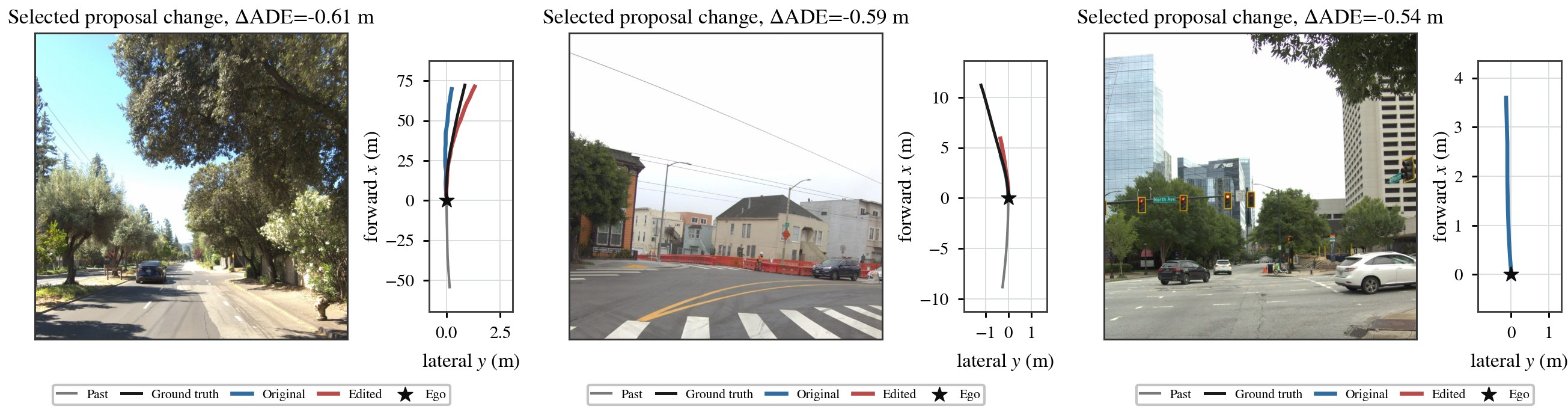}
    \caption{
\textbf{SAE-based scorer correction improves trajectory selection.}
The scoring adapter modifies proposal scores using SAE latents while leaving the base planner and generated proposal set fixed. Each example shows a case where score correction changes the selected trajectory to a lower-error proposal. \textit{Left:} The adapted scorer selects a trajectory that turns further into the side lane rather than continuing straight. \textit{Center:} The adapted scorer selects a higher-speed left-turn trajectory instead of a slower trajectory that cuts across the lane. \textit{Right:} The adapted scorer selects a stationary trajectory at a red light instead of accelerating into the intersection from a turn lane.
}
    \label{fig:finetune}
\end{figure}

Finally, we test whether the SAE representation is useful not only for interpretation, but also for improving the planner after training. Rather than fine-tuning the full \deepmonocularmodel, we freeze the base planner and train an adapter on top of its existing proposal-scoring submodule, as described in Section~\ref{sec:methods_33}. 

Adding the SAE-based adapter reduces ADE@5s by 3.44\% relative to the base model. Equivalently, the mean selected rank improves from 16.7 to 15.6 out of 50 proposal trajectories, where proposals are ranked by ADE to the ground-truth trajectory. Figure~\ref{fig:finetune} shows representative examples where score correction changes the selected trajectory, decreasing planning error.

As a control, we train the same adapter architecture without SAE latents, replacing them with the raw activations from the layer that the SAE is trained to reconstruct. This simplified control recovers 80\% of the ADE improvement achieved by the SAE-based correction adapter. In addition, we replace the frozen SAE encoder with a randomly initialized trainable encoder with the same architecture and parameter count. Despite being optimized directly for the downstream task, this control achieves only a 2.86\% ADE reduction. Together, these controls show that ADE improvement is not explained solely by adding adapter capacity on top of the scorer. Instead, the sparse SAE representation appears to provide a more useful interface for score correction, making relevant information more accessible than the dense query activation alone.

\FloatBarrier

\section{Conclusion}
We applied mechanistic interpretability to E2E autonomous driving by training Sparse
Autoencoders on activations of SMTS, a lightweight single-camera
trajectory scoring planner. SAE latents are structured, visually grounded, and steerable,
recovering route intent with 98.3\% accuracy unsupervised, and an SAE-based scoring
adapter reduces ADE@5s by 3.44\%, confirming their practical utility.

\textbf{Limitations --}
Our evaluation is entirely open-loop, so it does not measure how SAE features or
interventions behave under closed-loop feedback. SMTS uses only a front-facing camera,
leaving multi-camera or LiDAR-based planners unexplored. The counterfactual edit
analysis depends on the quality of the object detector and image editing model, and
imperfect edits may confound feature attribution. Finally, visual-edit responses and
latent steering effects are positively aligned but moderate in magnitude, indicating
that some visually responsive features do not translate cleanly into braking behavior.

\clearpage
\acknowledgments{This work used the Delta and DeltaAI systems at the National Center for Supercomputing Applications through allocation CIS260574 from the Advanced Cyberinfrastructure Coordination Ecosystem: Services \& Support (ACCESS) program, which is supported by U.S. National Science Foundation grants 2138259, 2138286, 2138307, 2137603, and 2138296 \citep{boerner2023access}.}

\bibliography{example}  %

\appendix

\section{Implementation Details}
  \label{sec:implementation-details}

  \paragraph{Dataset and input representation.}
  We use version 1.0.0 of the Waymo Open Dataset End-to-End Driving
  (WOD-E2E) camera dataset \cite{xu2025wode2ewaymoopendataset}. The complete indexed training and validation
  splits contain 415,663 and 106,360 samples, respectively. We train our end-to-end planning model (labeled SMTS)
  on 250,000 samples from the training split. The SAEs are trained
  using activations extracted from the complete training split.

  SMTS uses only the front-facing camera image. We decode each JPEG image
  to an RGB tensor, convert its values from uint8 to $[0,1]$, and apply
  the evaluation preprocessing associated with the pretrained
  \texttt{timm/vit\_pe\_spatial\_small\_patch16\_512.fb} backbone \cite{bolya2025PerceptionEncoder}. This
  consists of bicubic resizing and center cropping to $512 \times 512$,
  followed by channel-wise normalization with
  $\boldsymbol{\mu}=(0.5,0.5,0.5)$ and
  $\boldsymbol{\sigma}=(0.5,0.5,0.5)$. We do not apply deterministic or 
  stochastic image augmentation.

  Each sample contains a past ego state tensor
  $\mathbf{p}\in\mathbb{R}^{16\times6}$, where the six state variables are
  $(x,y,v_x,v_y,a_x,a_y)$. The 16 states are sampled at $4$\,Hz and
  represent four seconds of history. Route intent is represented as a
  three-dimensional one-hot vector corresponding to
  \{\textsc{straight}, \textsc{left}, \textsc{right}\}. The flattened
  96-dimensional state history and three-dimensional intent vector are
  concatenated to form the planner conditioning input. The model predicts
  $T=20$ future positions at $4$\,Hz, corresponding to a five-second
  prediction horizon, and generates $K=50$ trajectory proposals.

  \paragraph{SMTS architecture and training.}  We use the pretrained \\ \texttt{timm/vit\_pe\_spatial\_small\_patch16\_512.fb} vision
  transformer as the visual backbone \cite{bolya2025PerceptionEncoder}. The backbone is frozen throughout
  training, while the visual adapter and planner are trainable. We use the
  backbone's final spatial feature map, which contains $32\times32$ visual
  tokens with feature dimension $d=384$.

  The planner contains four transformer blocks. Each block updates one
  planner-query token using eight-head self-attention, eight-head
  cross-attention over the visual tokens, and an MLP with hidden dimension
  $4d$. The final query token is decoded into $K=50$ sequences of raw acceleration
  and angular-velocity controls. For proposal $k$ and timestep $t$, the raw
  controls $(\tilde{a}_{kt},\tilde{\omega}_{kt})$ are bounded as
  \begin{equation}
  a_{kt} = a_{\max}\tanh(\tilde{a}_{kt}),
  \qquad
  \omega_{kt} = \omega_{\max}\tanh(\tilde{\omega}_{kt}),
  \end{equation}
  where $a_{\max}=8.0\,\mathrm{m/s^2}$ and
  $\omega_{\max}=1.0\,\mathrm{rad/s}$. The initial position, speed, and
  heading are obtained from the final observed ego state:
  \begin{equation}
  x_{k0}=x_{\mathrm{past}},\quad
  y_{k0}=y_{\mathrm{past}},\quad
  v_{k0}=\sqrt{v_{x,\mathrm{past}}^2+v_{y,\mathrm{past}}^2+\varepsilon_v},
  \quad
  \theta_{k0}=\operatorname{atan2}(v_{y,\mathrm{past}},v_{x,\mathrm{past}}),
  \end{equation}
  with $\varepsilon_v=10^{-6}$. We then apply the differentiable kinematic rollout
  
  \begin{align}
  x_{k,t+1} &= x_{kt} + v_{kt}\cos(\theta_{kt})\Delta t, \\
  y_{k,t+1} &= y_{kt} + v_{kt}\sin(\theta_{kt})\Delta t, \\
  \theta_{k,t+1} &= \theta_{kt} + \omega_{kt}\Delta t, \\
  v_{k,t+1} &= \max\!\left(v_{kt}+a_{kt}\Delta t,\,0\right),
  \end{align}
  using $\Delta t=0.25$\,s for $T=20$ steps. Thus, each proposal represents
  a 5-second trajectory. More precisely, the implementation directly
  predicts yaw rate rather than steering angle and therefore does not
  explicitly model wheelbase.

  For proposal $k$ of sample $i$, we define its average displacement error
  as
  \begin{equation}
  e_{ik} =
  \frac{1}{T}\sum_{t=1}^{T}
  \left\lVert\hat{\mathbf{y}}_{ikt}-\mathbf{y}_{it}\right\rVert_2 .
  \end{equation}
  The trajectory loss is a top-5 winner-take-all objective,
  \begin{equation}
  \mathcal{L}_{\mathrm{traj}} =
  \frac{1}{5B}\sum_{i=1}^{B}\sum_{k\in\mathcal{I}_5(i)}e_{ik},
  \end{equation}
  where $\mathcal{I}_5(i)$ contains the five proposals with the lowest ADE.
  The proposal scorer predicts the ADE of each proposal and is trained
  using detached regression targets:
  \begin{equation}
  \mathcal{L}_{\mathrm{score}} =
  \frac{1}{BK}\sum_{i=1}^{B}\sum_{k=1}^{K}
  \left(s_{ik}-\operatorname{detach}(e_{ik})\right)^2 .
  \end{equation}
  Detaching the ADE targets and proposal trajectories prevents the scoring
  loss from modifying the trajectory generator.

  \paragraph{Adversarial scorer supervision.}
  To improve scorer robustness, we generate adversarial trajectories
  directly in Cartesian trajectory space during training. Starting from
  the detached proposal $\tilde{\mathbf{y}}_{ik}^{(0)}
  =\operatorname{detach}(\hat{\mathbf{y}}_{ik})$, we perform three
  projected gradient-ascent steps:
  \begin{equation}
  \tilde{\mathbf{y}}_{ik}^{(m+1)}
  =
  \Pi_{\mathcal{B}_{\infty}(\hat{\mathbf{y}}_{ik},\epsilon)}
  \left[
  \tilde{\mathbf{y}}_{ik}^{(m)}
  +
  \alpha\,
  \operatorname{sign}\!\left(
  \nabla_{\tilde{\mathbf{y}}_{ik}^{(m)}}s_{ik}^{(m)}
  \right)
  \right],
  \end{equation}
  where $\epsilon=0.1$, $\alpha=\epsilon/3$, and
  $\Pi_{\mathcal{B}_{\infty}}$ clips every trajectory coordinate to remain
  within $\epsilon$ of the original proposal. Gradient ascent increases
  the scorer's predicted ADE, producing locally difficult trajectories that
  improve the generalization of the scoring module to trajectories not 
  present in the original training dataset.
  We recompute the true ADE $\tilde e_{ik}$ of each perturbed trajectory
  and train the scorer to predict it:
  \begin{equation}
  \mathcal{L}_{\mathrm{adv}} =
  \frac{1}{BK}\sum_{i=1}^{B}\sum_{k=1}^{K}
  \left(\tilde{s}_{ik}-\operatorname{detach}(\tilde e_{ik})\right)^2 .
  \end{equation}

  \paragraph{Depth prediction head.}
  The model additionally contains an auxiliary depth head operating on
  the adapted visual feature map. The head uses two convolutional layers
  with GELU activations and nearest-neighbor upsampling, followed by a
  convolution and softplus activation, producing a
  $128\times128$ depth map. A frozen \texttt{Depth-Anything-V2-Small} model \cite{yang2024depthv2} generates
  pseudo-depth targets from the front image, and the auxiliary objective is
  \begin{equation}
  \mathcal{L}_{\mathrm{depth}}
  =
  \frac{1}{BHW}\sum_{i=1}^{B}
  \left\lVert
  \hat{\mathbf{D}}_i-\operatorname{detach}(\mathbf{D}^{*}_i)
  \right\rVert_1 .
  \end{equation}

  We train SMTS for 10 epochs using Adam with learning rate $10^{-4}$, no learning rate scheduling, and a global batch size of 128 on 4$\times$ NVIDIA A100 GPUs. The reported
  checkpoint uses
  \begin{equation}
  \mathcal{L}_{\mathrm{SMTS}}
  =
  \mathcal{L}_{\mathrm{traj}}
  +\mathcal{L}_{\mathrm{score}}
  +0.1\,\mathcal{L}_{\mathrm{adv}}
  +0.1\,\mathcal{L}_{\mathrm{depth}}
  \end{equation}

  \paragraph{Sparse autoencoders.}
  We independently train one sparse autoencoder (SAE) on the
  post-transformer query activations from each SMTS block
  $\ell\in\{0,1,2,3\}$. Each frame contributes one
  384-dimensional activation token per block. Consequently, each SAE is
  trained using 415,663 training tokens and evaluated using 106,360
  validation tokens.

  Each SAE has a latent dimension of 384. The encoder uses Top-$k$ sparsification with $k=24$ and no
  encoder bias. The AuxK objective uses $k_{\mathrm{aux}}=512$, restricted
  to currently dead latents, with coefficient
  $\alpha_{\mathrm{aux}}=1/32$. A latent is considered dead when it has
  not activated for 500 consecutive optimizer steps.

  Before encoding, we subtract a learned preprocessing bias initialized
  using an approximate geometric median of 10,000 training tokens. For
  each activation, we then subtract its mean across feature dimensions and
  divide by its centered $L_2$ norm. Reconstruction loss is normalized
  by the centered mean-squared activation magnitude. Decoder columns are
  renormalized to unit norm after every optimizer update.

  The SAE objective is
  \begin{equation}
  \mathcal{L}_{\mathrm{SAE}} =
  \frac{\operatorname{MSE}(\hat{\mathbf{x}},\mathbf{x})}
       {\operatorname{CMSE}(\mathbf{x})}
  +
  \frac{1}{32}
  \frac{\operatorname{MSE}(\hat{\mathbf{r}}_{\mathrm{aux}},
                           \mathbf{r})}
       {\operatorname{CMSE}(\mathbf{x})},
  \end{equation}
  where $\mathbf{r}$ is the reconstruction residual and
  $\hat{\mathbf{r}}_{\mathrm{aux}}$ is reconstructed using dead latents.

  We train each SAE for 100 epochs using Adam with learning rate
  $10^{-4}$, $(\beta_1,\beta_2)=(0.9,0.999)$, batch size 4096, gradient
  clipping at norm 1.0, and random seed 42. We retain the checkpoint with
  the lowest validation loss.

  \FloatBarrier

\section{Open-Loop Planning Performance}
\label{sec:app_open_loop}

We compare against Latent TransFuser \citep{chitta2022transfuserimitationtransformerbasedsensor},
GTRS-Dense \citep{li2025generalizedtrajectoryscoringendtoend}, and DrivoR
\citep{kirby2026drivingregisters}. These models were developed for other
datasets, so we re-train DrivoR and GTRS-Dense on the same 250,000-sample subset. These models were chosen for comparison to \deepmonocularmodel\ as they are closest to \deepmonocularmodel\ in design and parameter count.

\begin{figure}[htpb]
    \centering
    \begin{tikzpicture}
    \begin{axis}[
        width=0.78\linewidth,
        height=0.50\linewidth,
        xlabel={Training samples ($\times 10^3$)},
        ylabel={ADE@5s},
        xmin=45, xmax=255,
        ymin=1.35, ymax=1.85,
        xtick={50,100,150,200,250},
        grid=both,
        grid style={line width=.1pt, draw=gray!20},
        major grid style={line width=.2pt, draw=gray!45},
        legend style={
            at={(0.98,0.98)},
            anchor=north east,
            draw=none,
            fill=white,
            fill opacity=0.75,
            text opacity=1
        },
        tick label style={font=\small},
        label style={font=\small},
        legend style={font=\small},
    ]

    \addplot[
        thick,
        loosely dotted,
        color=blue!70!black,
        mark=*,
    ] coordinates {
        (50, 1.589482)
        (100, 1.588190)
        (150, 1.520209)
        (200, 1.538604)
        (250, 1.559245)
    };
    \addlegendentry{GTRS-Dense}

    \addplot[
        thick,
        loosely dotted,
        color=orange!85!black,
        mark=square*,
    ] coordinates {
        (50, 1.811242)
        (100, 1.630712)
        (150, 1.645586)
        (200, 1.620093)
        (250, 1.628125)
    };
    \addlegendentry{DrivoR}

    \addplot[
        thick,
        densely dotted,
        color=green!55!black,
        mark=triangle*,
    ] coordinates {
        (50, 1.579329)
        (100, 1.473576)
        (150, 1.477759)
        (200, 1.425356)
        (250, 1.400000)
    };
    \addlegendentry{SMTS}

    \end{axis}
    \end{tikzpicture}
    \caption{
\textbf{Data scaling.} Selected trajectory ADE@5s on the WOD-E2E validation split as a function of training set size. \deepmonocularmodel \ shows more consistent gains from additional training data than the compared trajectory scoring baselines and achieves the lowest error overall.
}
    \label{fig:ade5_vs_train_size}
\end{figure}
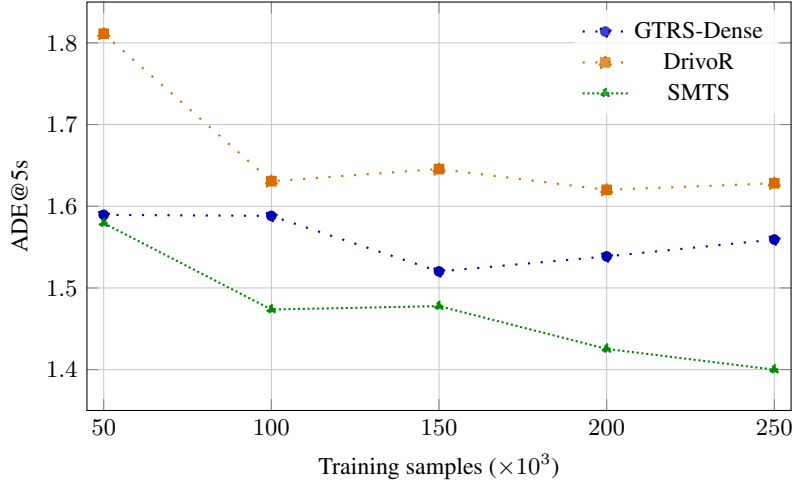

\FloatBarrier

\begin{table}[htpb]
\centering
\small
\setlength{\tabcolsep}{5pt}
\begin{tabular}{lccc}
\toprule
Scene subset & GTRS-Dense & DrivoR & \deepmonocularmodel\ (ours) \\
\midrule
\textbf{Average}      & 7.688          & 7.645          & \textbf{7.765} \\
\midrule
Construction          & 8.080          & 8.147          & \textbf{8.343} \\
Intersection          & \textbf{7.948} & 7.632          & 7.706          \\
Pedestrian            & 7.782          & 7.777          & \textbf{7.977} \\
Cyclist               & 7.613          & \textbf{7.636} & 7.569          \\
Multi-lane maneuver   & 7.409          & 7.483          & \textbf{7.614} \\
Single-lane maneuver  & 8.194          & 8.037          & \textbf{8.299} \\
Cut-in                & 7.821          & \textbf{7.867} & 7.715          \\
Foreign object debris & 7.891          & 7.860          & \textbf{7.988} \\
Special vehicle       & 8.030          & 7.687          & \textbf{8.078} \\
Spotlight             & 6.295          & \textbf{6.773} & 6.749          \\
Others                & \textbf{7.499} & 7.193          & 7.372          \\
\bottomrule
\end{tabular}
\vspace{2mm}
\caption{\textbf{WOD-E2E test-set Rater Feedback Score by scene subset}
(higher is better). All models are re-trained on the same 250,000-sample
training subset and scored by the private WOD-E2E test set evaluator.}
\label{tab:rfs_scores}
\end{table}

Figure~\ref{fig:ade5_vs_train_size} shows that \deepmonocularmodel \ benefits more consistently from additional training data than the compared baselines. While \deepmonocularmodel \ performs similarly to GTRS-Dense at 50,000 training samples, its ADE decreases more consistently as the dataset size increases, ultimately achieving the lowest ADE at 250,000 samples. Furthermore, \deepmonocularmodel\ obtains the highest average RFS and leads on 6 of 11 scene
subsets (Table~\ref{tab:rfs_scores}), with the largest margins on construction,
pedestrian, and multi-lane maneuver scenes.

\section{Feature Clustering Details}
\label{sec:supp-feature-clustering}

We analyze co-activation structure among the block-3 SAE features on all
\(N=106{,}360\) WOD-E2E validation scenes. We use block~3 because it is the
final planner representation immediately preceding trajectory generation and
proposal scoring.

\subsection{Activation Filtering}
\label{sec:supp-activation-filtering}

Let \(z_{ij}\geq0\) denote the activation of SAE feature \(j\) on validation
scene \(i\). We define the binary firing indicator and activation frequency as
\begin{equation}
    b_{ij}=\mathbbm{1}[z_{ij}>0],
    \qquad
    f_j=\frac{1}{N}\sum_{i=1}^{N}b_{ij}.
\end{equation}
We retain features satisfying
\begin{equation}
    0.005 \leq f_j \leq 0.50.
\end{equation}
The lower bound removes features that fire too rarely to support reliable
correlation estimates, while the upper bound removes features that fire for many scenes
and do not provide information specific to a scene.

Of the \(384\) block 3 SAE features, \(258\) are retained after filtering, as
\(125\) features with activation frequencies below \(0.5\%\) and one feature
with an activation frequency above \(50\%\) are excluded. Among the retained features, the
observed activation frequencies range from \(0.509\%\) to \(46.542\%\).

For every retained feature pair \((j,\ell)\), we compute the Pearson
correlation between their binary firing indicators:
\begin{equation}
    \rho_{j\ell}
    =
    \operatorname{corr}
    \left(
        \mathbf{b}_{:j},
        \mathbf{b}_{:\ell}
    \right).
\end{equation}
For binary variables, this correlation is equivalent to the phi coefficient
and measures whether two features fire on the same scenes more frequently
than expected from their marginal firing rates.

We construct an undirected graph in which retained features are nodes and an
edge connects \(j\) and \(\ell\) when
\begin{equation}
    \rho_{j\ell}\geq\tau.
\end{equation}
Clusters are the graph's connected components, and components containing
fewer than three features are discarded. Because connected components use
transitive connectivity, not every feature pair within a cluster is required
to have correlation above \(\tau\).

\subsection{Robustness of the Clustering Procedure}
\label{sec:supp-cluster-threshold}

We evaluate correlation thresholds
\(\tau\in\{0.20,0.35,0.40\}\) and use \(\tau=0.35\) for the primary analysis.
Table~\ref{tab:cluster-threshold-sensitivity} summarizes the resulting
cluster structures.

\begin{table*}[t]
    \centering
    \small
    \caption{
        Sensitivity of block 3 SAE feature clusters to the co-activation
        correlation threshold. Cluster sizes are reported in descending
        order. Intent purity is measured among each cluster's top 50
        validation scenes by activation.
    }
    \label{tab:cluster-threshold-sensitivity}
    \begin{tabular}{c c l c c}
        \toprule
        Threshold \(\tau\) &
        Number of clusters &
        Cluster sizes &
        Clustered features &
        Top 50 intent purity \\
        \midrule
        \(0.20\) & 1 & \(238\) & \(238\) & \(100\%\), collapsed component \\
        \(0.35\) & 4 & \(64,36,34,27\) & \(161\) & \(100\%\) for all clusters \\
        \(0.40\) & 6 & \(43,29,27,23,3,3\) & \(128\) & \(96\%\)--\(100\%\) \\
        \bottomrule
    \end{tabular}
\end{table*}

At \(\tau=0.20\), weak correlation paths transitively connect \(238\) of the
\(258\) retained features into one large component. Although its highest
activating scenes all have straight intent, this component combines otherwise
distinct driving modes and is therefore too coarse for interpretation.

The four primary clusters identified at \(\tau=0.35\) remain present at
\(\tau=0.40\). The stopped/slow-straight, stopped left turn,
high speed straight, and stopped right turn clusters retain respectively
\(43/64\), \(27/36\), \(29/34\), and \(23/27\) of their features. Their
dominant intents remain unchanged, and each retains \(100\%\) top-50 intent
purity. Intent purity is the fraction of a cluster's top 50 validation
scenes, by activation, that share the most common intent label. For example, a purity
of \(100\%\) with dominant intent ``Left'' means that all 50 scenes are
labeled \textsc{left}. The two additional components at \(\tau=0.40\) contain only three
features each. Thus, the principal intent-aligned structure is stable across
the two stricter thresholds, while \(\tau=0.35\) preserves more features
within each semantic mode.

As an additional robustness check, we cluster the retained features using
greedy modularity maximization, which does not require a correlation threshold
\citep{clauset2004finding,newman2004finding}. After matching each recovered
community to the connected-component cluster with which it has the greatest
overlap, 94.3\% of retained features receive the same label under both
procedures. The dominant cluster structure is therefore largely preserved
under an alternative clustering method.

\subsection{Cluster-Level Scene Summaries}
\label{sec:supp-cluster-scenes}

For a cluster \(\mathcal{C}\), we define its activation score on scene \(i\)
using the sum of its raw SAE activation magnitudes:
\begin{equation}
    A_i(\mathcal{C})
    =
    \sum_{j\in\mathcal{C}} z_{ij}.
\end{equation}
We rank validation scenes by \(A_i(\mathcal{C})\) and summarize each cluster
using its top \(50\) scenes. This ranking uses activation magnitude only after
clusters have been constructed from binary co-activation correlations.

Table~\ref{tab:cluster-scene-summaries} reports the four primary clusters at
\(\tau=0.35\). Continuous scene statistics are standardized relative to their
distribution over the complete validation split.

\begin{table*}[t]
    \centering
    \small
    \caption{
        Block-3 SAE cluster summaries at the primary threshold
        \(\tau=0.35\). Intent and standardized scene statistics are computed
        over each cluster's top 50 validation scenes by activation.
    }
    \label{tab:cluster-scene-summaries}
    \begin{tabular}{c c c l r r r}
        \toprule
        Cluster &
        Features &
        Top-50 intent &
        Interpretation &
        Speed \(z\) &
        Braking \(z\) &
        Score entropy \(z\) \\
        \midrule
        C0 & 64 & Straight (\(100\%\)) & Slow straight driving
           & \(-1.04\) & \(-0.52\) & \(+0.03\) \\
        C1 & 36 & Left (\(100\%\)) & Slow left turn
           & \(-1.07\) & \(+2.00\) & \(-1.75\) \\
        C2 & 34 & Straight (\(100\%\)) & High-speed straight driving
           & \(+0.98\) & \(-0.30\) & \(+0.48\) \\
        C3 & 27 & Right (\(100\%\)) & Slow right turn
           & \(-1.04\) & \(+1.12\) & \(-0.94\) \\
        \bottomrule
    \end{tabular}
\end{table*}

The clusters separate both route intent and dynamic driving state. C1 and C3
distinguish left- and right-turn intent despite both primarily activating in
slow or stationary scenes. C0 and C2 share straight intent but separate
stationary and high-speed motion. These results show that the final
planner representation is organized into distributed feature groups encoding
combinations of route intent and ego motion.

\begin{figure*}[t]
    \centering
    \begin{subfigure}[t]{0.24\textwidth}
        \centering
        \includegraphics[width=\linewidth]
        {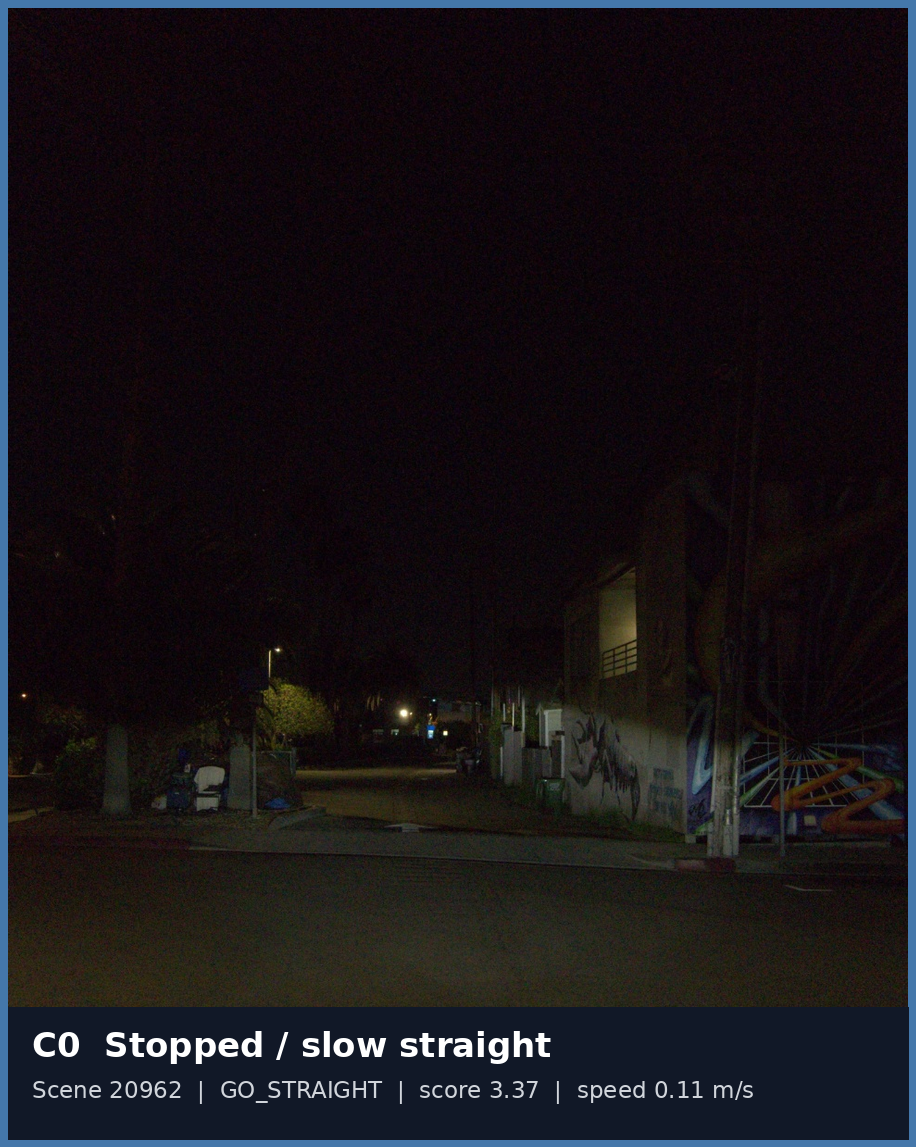}
    \end{subfigure}
    \hfill
    \begin{subfigure}[t]{0.24\textwidth}
        \centering
        \includegraphics[width=\linewidth]
        {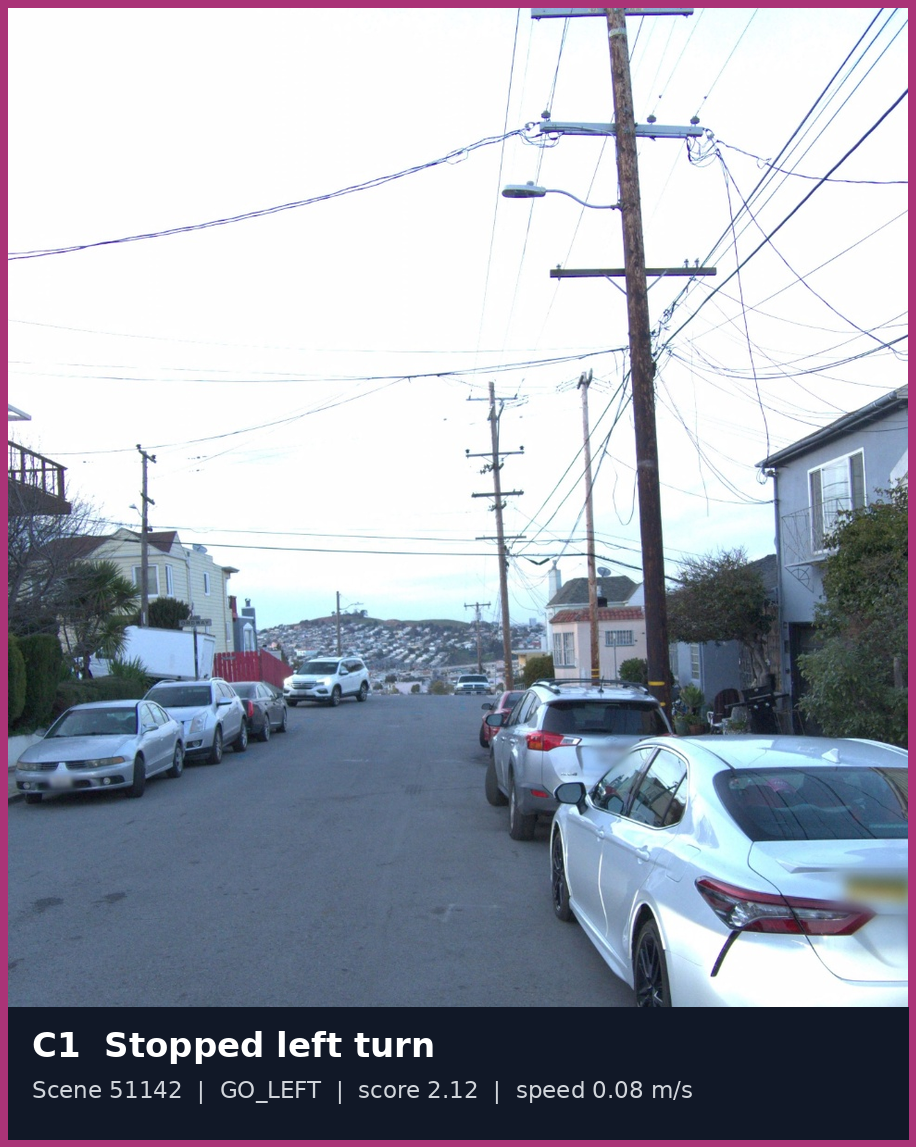}
    \end{subfigure}
    \hfill
    \begin{subfigure}[t]{0.24\textwidth}
        \centering
        \includegraphics[width=\linewidth]
        {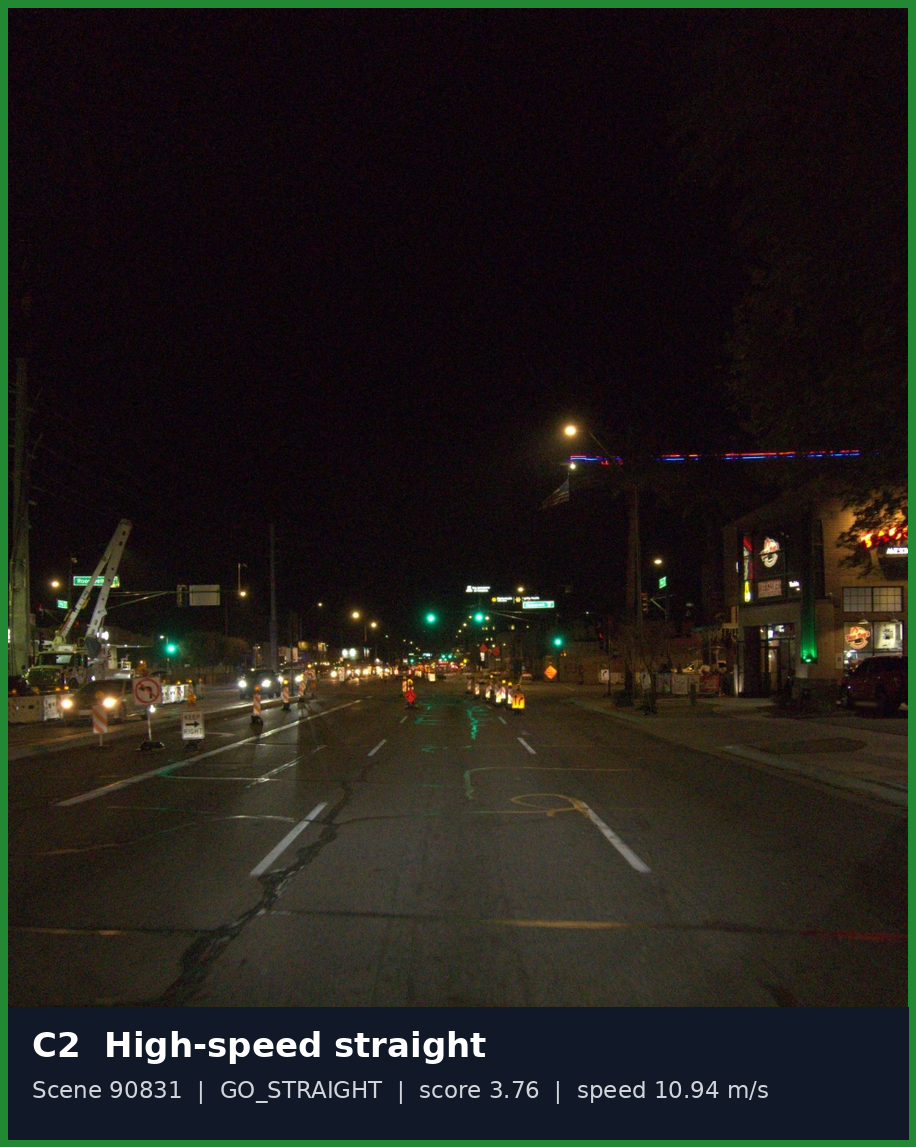}
    \end{subfigure}
    \hfill
    \begin{subfigure}[t]{0.24\textwidth}
        \centering
        \includegraphics[width=\linewidth]
        {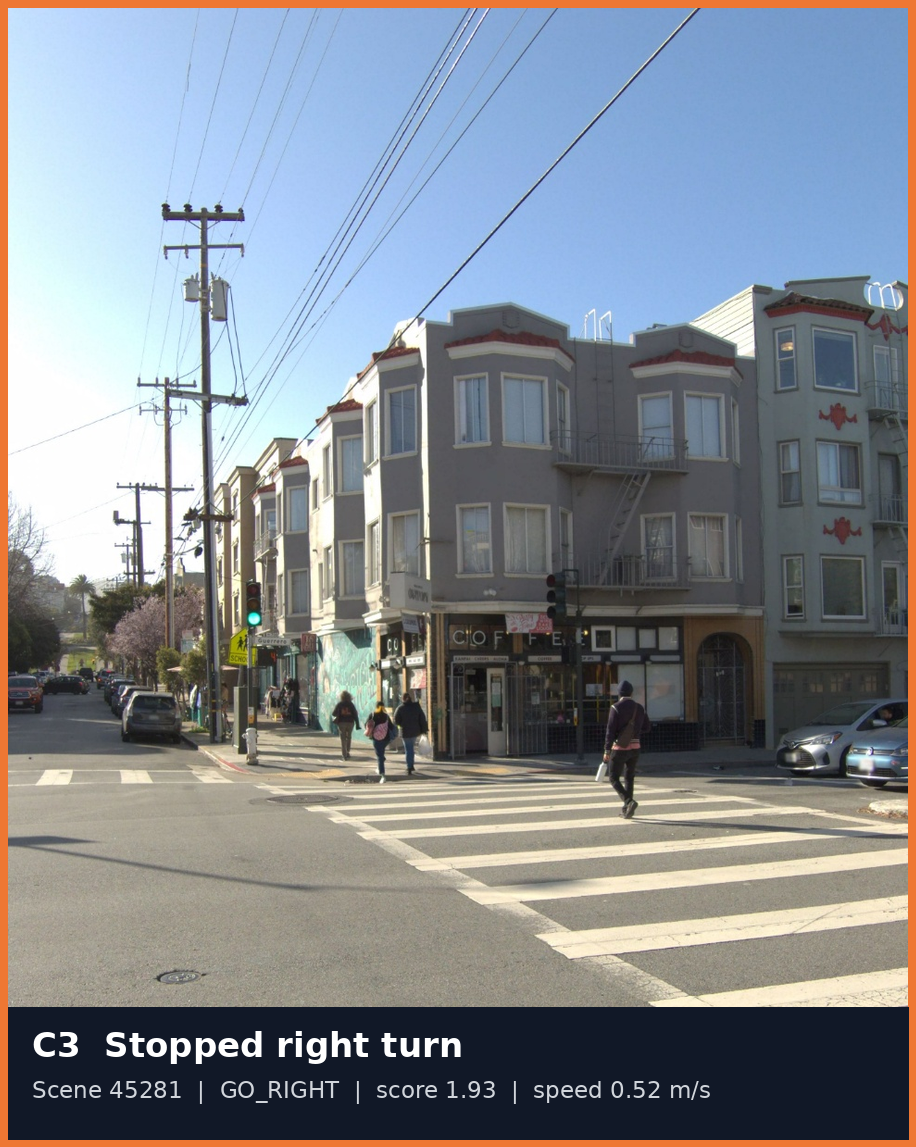}
    \end{subfigure}

    \caption{
        Highest activating validation scene for each feature cluster.
        The clusters capture distinct combinations of route intent and
        vehicle motion.
    }
    \label{fig:cluster-top-scenes}
\end{figure*}

\section{Input Conditioning Analysis}
\label{sec:supp-variance-decomposition}

We test whether SAE latents depend on visual input beyond the
route intent and ego history used to initialize the query. We sample
9{,}000 WOD-E2E validation frames, balanced across the six combinations of
original route intent
$\{\textsc{straight},\textsc{left},\textsc{right}\}$
and whether the vehicle is moving or stopped, with 1{,}500 frames from each
group. This balancing prevents common driving conditions from dominating the
analysis.

For each sampled image, we keep the image unchanged and rerun the planner
with each of the three route intents and two standardized ego histories. The
first history represents a stopped vehicle, with position, velocity, and
acceleration set to zero. The second represents steady forward motion at
$5.96$~m/s, with zero acceleration and positions advanced at $0.25$~s
intervals. The cruise speed of $5.96$~m/s was chosen as the median speed over all moving (where moving is defined as having a speed greater than 0.5 m/s) validation frames.
Each image is therefore evaluated under the same six combinations of intent
and ego history, producing $54{,}000$ latent vectors.

We then use a three-way ANOVA to separate variation in the latent
representation associated with image, route intent, ego history, and their
interactions. For each latent dimension, we compute the sum of
squares associated with each term, sum these values across dimensions, and
normalize by the total sum of squares. The resulting percentages are reported
in Table~\ref{tab:variance-decomposition}.

The main effect of image measures how much the latent representation differs
across images after averaging over intent and ego history. The
image$\times$intent term measures how much the effect of route intent differs
from image to image, with the remaining interaction terms interpreted
similarly. Because every image is evaluated under every intent and ego
history combination, these effects are not caused by differences in the
metadata paired with each image.

  \section{Counterfactual Visual Edit Details}
  \label{sec:supp-edits}
  This section details the counterfactual image-editing pipeline we use
  to examine the visual responsiveness of SAE features for our end-to-end
  driving model.
  \subsection{Edit generation pipeline}
  \label{sec:supp-edit-pipeline}
  For each scene we operate on the front camera image and apply a four-stage
  pipeline:
  \begin{enumerate}
    \item \textbf{Object detection.} We run a YOLO detector
          (\texttt{yolo26x}) on the front image to obtain a structured scene
          description (object classes, confidences, and bounding boxes) \cite{sapkota2026yolo26keyarchitecturalenhancements}.
    \item \textbf{Edit proposal (VLM).} The scene description and image are passed to a Qwen3.5-4B vision–language model (VLM) \cite{qwen35blog}. The VLM is constrained by a fixed
          system prompt to propose at most one counterfactual
          edit and to return JSON with the schema
          (\texttt{edit\_type}, \texttt{edit\_direction}, \texttt{prompt}).
    \item \textbf{Allowed edits.} Proposals are validated against a closed
          vocabulary. The permitted edit types are traffic light color,
          traffic light presence, pedestrian presence, and stop sign
          presence or location; the permitted directions are
          \texttt{green\_to\_red}, \texttt{red\_to\_green},
          \texttt{yellow\_to\_green}, \texttt{add\_traffic\_light},
          \texttt{remove\_traffic\_light}, \texttt{add\_pedestrian},
          \texttt{remove\_pedestrian}, \texttt{add\_stop\_sign},
          \texttt{remove\_stop\_sign}, and \texttt{move\_stop\_sign}.
          Any scene with no applicable edit, or with an unparseable/invalid
          proposal, is skipped.
    \item \textbf{Image editing.} Valid prompts are rendered by an
          instruction-guided diffusion image editor (\texttt{FireRed-Image-Edit}) \cite{superintelligenceteam2026fireredimageedit10technicalreport}.
          We manually engineered the image generation system prompt to ensure instruction
          following and minimize unrelated changes to the image.
  \end{enumerate}
  \paragraph{Replay protocol.}
  To measure the effect of an edit, we substitute the edited image for
  the original front camera frame, keep all other inputs (past state and intent) fixed, and rerun the planner. We extract the block 3
  query token for both the original and edited inputs, encode both with
  the SAE, and record the per-feature activation change
  $\Delta z_i = z_i^{\text{edited}} - z_i^{\text{original}}$, which we then
  normalize across the validation split. This isolates how a targeted visual change
  propagates into the SAE feature space and, in turn, the planner output.
  \subsection{Edit category counts}
  \label{sec:supp-edit-counts}
  We processed $1000$ validation scenes. The VLM proposed an applicable edit
  for 615 of them and returned \texttt{no\_change} for the
  remaining 385 (21 of which were rejected for invalid edit
  specifications). 
  \begin{table}[h]
  \centering
  \begin{tabular}{lr}
  \toprule
  Edit direction & Count \\
  \midrule
  green $\rightarrow$ red & 193 \\
  red $\rightarrow$ green & 95 \\
  add stop sign & 46 \\
  add pedestrian & 145 \\
  \midrule
  no change / rejected & 385 \\
  \bottomrule
  \end{tabular}
  \vspace{2mm}
  \caption{Counterfactual visual edit dataset summary.
  The four directions above are those used in the main
  edit sensitivity analysis. A further $136$ edits in other categories were
  also generated (\texttt{add\_traffic\_light} $46$, \texttt{yellow\_to\_green}
  $41$, \texttt{remove\_stop\_sign} $25$, \texttt{remove\_pedestrian} $14$,
  \texttt{remove\_traffic\_light} $7$, \texttt{move\_stop\_sign} $3$), resulting in
  $615$ edited images in total.}
  \end{table}
  \subsection{Additional examples}
  \label{sec:supp-edit-examples}
\begin{figure}[t]
  \centering
  \includegraphics[width=\linewidth]{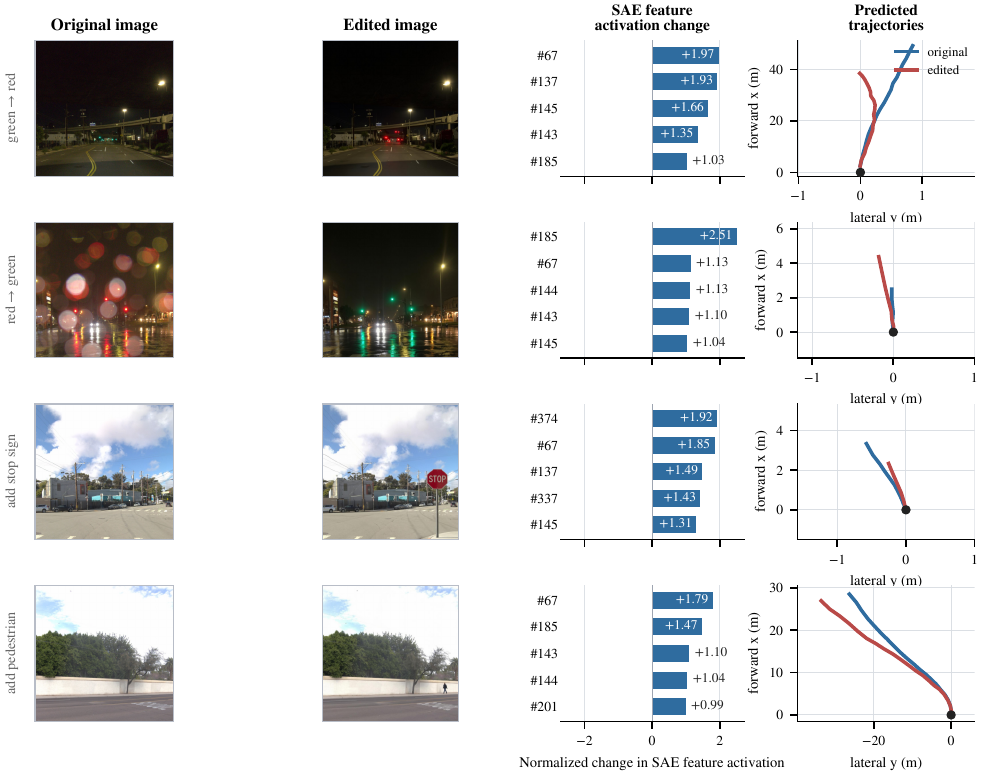}
  \caption{Additional counterfactual visual edit examples.
  Each row shows the original front camera image, the edited image, the top 5
  SAE feature activation changes for that edit direction (normalized
  $\Delta z_i$), and the planner's predicted trajectory before
  (blue) and after (red) the edit. Stop-inducing edits shorten or brake the
  predicted path; \texttt{red$\rightarrow$green} extends it.}
  \label{fig:supp-edit-examples}
  \end{figure}

  Figure~\ref{fig:supp-edit-examples} shows representative successful edits as
  original/edited image pairs. For each pair we overlay the planner's predicted
  trajectory before and after the edit and list the top SAE feature activation
  changes $\Delta z_i$. Stop-inducing edits (e.g.\ green$\rightarrow$red,
  add stop sign, add pedestrian) shorten or add braking to the predicted trajectory and
  increase the corresponding stop-associated SAE features.
  \subsection{Failure cases}
  \label{sec:supp-edit-failures}
\begin{figure}[t]
  \centering
  \includegraphics[width=\linewidth]{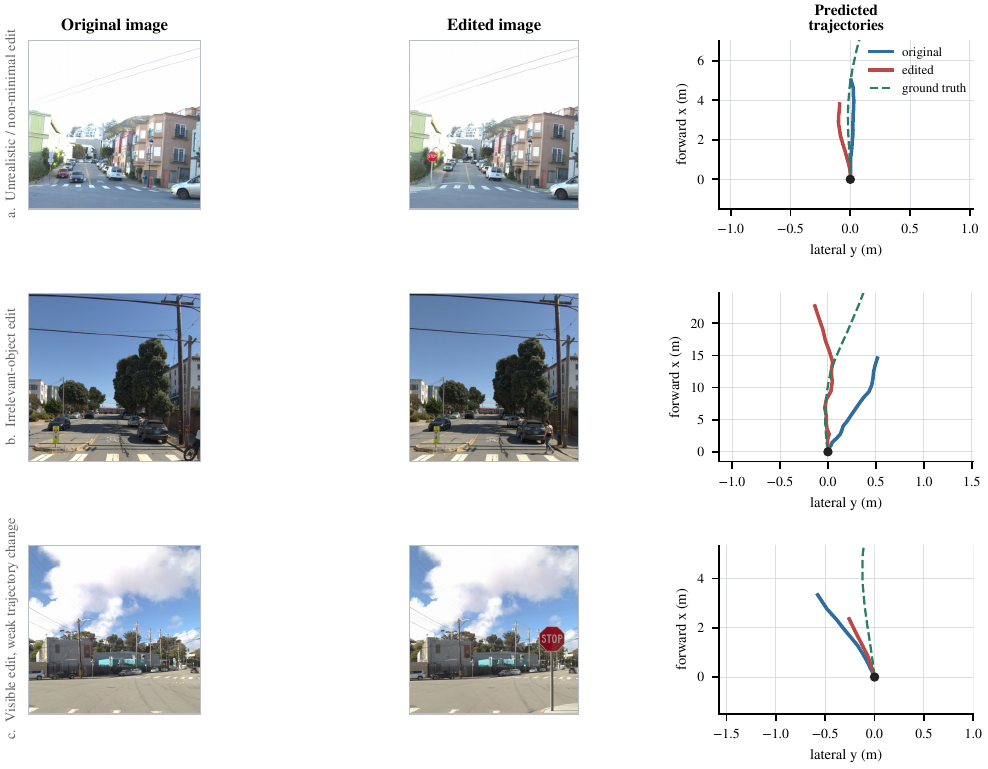}
  \caption{Representative failure modes of the counterfactual editing pipeline.
  \textbf{(a)}~Unrealistic edit: an oddly placed stop sign is added, and the
  editor warps the scene and deletes parked cars.
  \textbf{(b)}~Irrelevant object edit: a pedestrian is added on the right
  sidewalk, away from the intended driving path.
  \textbf{(c)}~Visible edit, weak trajectory change: a clear stop sign is added,
  yet the predicted path shortens only slightly relative to the original.
  Note that blue/red curves are the selected planner proposals before/after the edit and
  the dashed green curve is the ground truth future trajectory.}
  \label{fig:supp-edit-failures}
  \end{figure}

  For transparency, we also show failure modes of our pipeline
  (Figure~\ref{fig:supp-edit-failures}): (i)~\emph{unrealistic edits}, where the
  editor introduces visually implausible content; (ii)~\emph{irrelevant-object
  edits}, where the change is applied to an object that does not affect driving;
  and (iii)~\emph{behaviorally inert edits}, where a visible image change does
  not alter the predicted trajectory. These cases motivate our focus on
  stop-inducing edits, for which the visual-to-behavior link is strongest and
  most consistent.

\section{Latent Steering}
\label{sec:supp_latent_steering}

\subsection{Metrics} \label{sec:metrics}

Define $p_{k,\tau}\in\mathbb{R}^2$ to be
proposal $k$ at horizon step $\tau$, $s_k$ its score, and
$\hat{k}=\arg\min_k s_k$ the selected proposal. For the selected trajectory,
we define $\Delta p_\tau=p_{\hat{k},\tau}-p_{\hat{k},\tau-1}$,
$\phi_\tau=\operatorname{atan2}(\Delta p_{\tau,y},\Delta p_{\tau,x})$, and let
$a_\tau$ denote longitudinal control.

\begin{table}[!htp]
\centering
\small
\begin{tabular}{lll}
\toprule
Metric & Definition \\
\midrule
Final lateral displacement &
$d_{\mathrm{lat}} = p_{\hat{k},T,y}$ &  \\[4pt]

Average curvature &
$\displaystyle
\bar{\kappa}
=
\frac{1}{T-2}
\sum_{\tau=2}^{T-1}
\frac{|\phi_\tau-\phi_{\tau-1}|}
{\frac{1}{2}(\|\Delta p_\tau\|+\|\Delta p_{\tau-1}\|)+\epsilon}
$ &  \\[10pt]

Brake magnitude &
$\displaystyle
\beta =
\frac{1}{T}\sum_{\tau=1}^{T}\max(0,-a_\tau)
$ &  \\[8pt]

Acceleration magnitude &
$\displaystyle
\gamma =
\frac{1}{T}\sum_{\tau=1}^{T}\max(0,a_\tau)
$ &  \\[8pt]

Score margin &
$\displaystyle
m=s_{(2)}-s_{(1)}
$ & \\[8pt]

Proposal spread &
$\displaystyle
\sigma_{\mathrm{prop}}
=
\frac{1}{KT}
\sum_{k=1}^{K}
\sum_{\tau=1}^{T}
\|p_{k,\tau}-\bar{p}_\tau\|_2
$ & \\
\bottomrule
\end{tabular}
\caption{Metrics recorded at each intervention level.}
\label{tab:intervention_metrics}
\end{table}

\FloatBarrier

\subsection{Feature Quality} \label{sec:appendix_feature_quality}

The quality of an SAE feature as a controller is calculated through the following measures:

\paragraph{\textit{Standardized Response Magnitude.}}
For each SAE feature $i$ and target statistic $m$, we measure the change induced by the maximum intervention level, averaged over the scenes where feature $i$ is most active. Let $S_i$ denote this set of scenes, and let $y_m^{(s)}(\alpha)$ be the value of statistic $m$ for scene $s$ after intervening on feature $i$ with strength $\alpha$. We define the response per scene at the maximum intervention level as
\[
\Delta_{i,m}^{(s)}
=
y_m^{(s)}(\alpha_{\max}) - y_m^{(s)}(0).
\]
To make effects comparable across statistics with different units, we normalize by the standard deviation across the validation split, $\sigma_m$, of statistic $m$:
\[
E_{i,m}
=
\frac{
\frac{1}{|S_i|}\sum_{s \in S_i} \Delta_{i,m}^{(s)}
}{
\sigma_m
}.
\]
The absolute standardized response magnitude is then
\[
A_{i,m} = |E_{i,m}|.
\]

\paragraph{\textit{Monotonicity.}}
A feature intervention is more interpretable if increasing the intervention strength produces an ordered change in the target statistic. For each scene $s$, we compute the Spearman rank correlation between the intervention levels and the resulting statistic values:
\[
\rho_{i,m}^{(s)}
=
\operatorname{Spearman}
\left(
[\alpha_0, \ldots, \alpha_L],
[y_m^{(s)}(\alpha_0), \ldots, y_m^{(s)}(\alpha_L)]
\right).
\]
The feature-level monotonicity score is the average Spearman correlation across scenes:
\[
R_{i,m}
=
\frac{1}{|S_i|}
\sum_{s \in S_i}
\rho_{i,m}^{(s)}.
\]
Since both increasing and decreasing monotonic responses indicate controllability, the final controllability score uses $|R_{i,m}|$.

\paragraph{\textit{Response Consistency.}}
A large average response may be caused by a small number of outlier scenes. To penalize this, we define response consistency as the fraction of scenes whose response agrees with the average direction of effect. Let
\[
g_{i,m} = \operatorname{sign}(E_{i,m}).
\]
A scene is counted as consistent if its response curve is monotonic in the global direction and if its response to the maximal intervention has the same sign as the global response:
\[
g_{i,m}\rho_{i,m}^{(s)} > 0.5
\quad \text{and} \quad
g_{i,m}\Delta_{i,m}^{(s)} > 0.
\]
The response consistency score is
\[
C_{i,m}
=
\frac{1}{|S_i|}
\sum_{s \in S_i}
\mathbf{1}
\left[
g_{i,m}\rho_{i,m}^{(s)} > 0.5
\;\land\;
g_{i,m}\Delta_{i,m}^{(s)} > 0
\right].
\]

\paragraph{\textit{Response Selectivity.}}
We also require a controllable feature to affect the target statistic more strongly than the other measured planner statistics. Let $\mathcal{M}$ be the set of target statistics:
\[
\begin{aligned}
\mathcal{M} = \{ & \text{final lateral displacement, average curvature, brake magnitude,} \\
& \text{acceleration magnitude, score margin, proposal spread} \}
\end{aligned}
\]
For each feature-statistic pair, we define response selectivity as the share of the feature's total standardized effect assigned to statistic $m$:
\[
Q_{i,m}
=
\frac{
A_{i,m}
}{
\sum_{m' \in \mathcal{M}} A_{i,m'} + \epsilon
}.
\]
This term is large when feature $i$ affects statistic $m$ more selectively than the other measured statistics.

\paragraph{\textit{Controllability Score.}}
\label{sec:control_score}
Finally, we combine standardized response magnitude, monotonicity, response consistency, and response selectivity into a scalar controllability score:
\[
\operatorname{Ctrl}_{i,m}
=
A_{i,m}
\cdot
|R_{i,m}|
\cdot
C_{i,m}
\cdot
Q_{i,m}.
\]
Features are ranked separately for each statistic according to $\operatorname{Ctrl}_{i,m}$. 

We perform an exhaustive intervention sweep to test whether latent-steering
behavior generalizes beyond an individually selected
feature or scene. Unless otherwise stated, we intervene on the block-3 SAE
latents and evaluate the 64 validation scenes with the highest natural
activation for each feature. Of the 384 SAE features, 340 activate on at least
one validation scene.

\subsection{Intervention Protocol and Single-Feature Responses}
\label{sec:supp_single_feature_steering}

For feature $j$, we define its intervention scale as
\begin{equation}
s_j =
\max\left(
    \operatorname{Std}[z_j \mid z_j > 0],
    \frac{1}{4}\operatorname{Mean}[z_j \mid z_j > 0],
    0.05
\right).
\end{equation}
We increase the feature activation according to
\begin{equation}
z_j^{(\alpha)}
=
\max\left(0, z_j + \alpha s_j\right),
\qquad
\alpha \in \{0, 0.5, 1, 2\},
\end{equation}
while leaving all other SAE latents unchanged. The modified latent vector is
decoded back into the reconstructed query and passed through the remaining
planner heads. For each intervention, we measure the trajectory and scoring statistics defined
in Section \ref{sec:metrics}.

\subsection{Top Controllable Features}
\label{sec:supp_top_controllable}

We rank features using the controllability score defined in
Section~\ref{sec:appendix_feature_quality}. Table~\ref{tab:top_controllable_features} reports the strongest robust feature
for each statistic, restricting the ranking to features with consistency of at least $0.5$. 
In total, 254 of the 340 active
block 3 features obtain a control score above $0.01$, and 68 obtain a score
above $0.1$.

\begin{table*}[t]
\centering
\small
\begin{tabular}{lrrrrrr}
\toprule
Target statistic & Feature & Control score & Effect size &
Monotonicity & Consistency & Selectivity \\
\midrule
Brake magnitude       & 137 & 0.481 & $+0.715$ & 0.941 & 1.000 & 0.715 \\
Curvature             & 118 & 0.113 & $+0.472$ & 0.509 & 0.641 & 0.736 \\
Lateral displacement  & 326 & 0.330 & $+0.431$ & 0.947 & 0.953 & 0.848 \\
Score margin          & 228 & 1.440 & $+3.028$ & 0.959 & 0.828 & 0.599 \\
Proposal spread       & 250 & 0.410 & $+0.651$ & 1.000 & 1.000 & 0.629 \\
\bottomrule
\end{tabular}
\caption{Top robust block-3 SAE feature for each planner statistic. Effect
sizes are normalized by the validation-set standard deviation of the
corresponding statistic.}
\label{tab:top_controllable_features}
\end{table*}

The next highest robust braking features are 87 and 114; the next highest
lateral-displacement features are 207 and 265; and the next highest
proposal spread features are 153 and 116. Score margin feature 228 is the
strongest robust scorer-related feature. Features 137, 228, and 250 belong to
clusters C2, C0, and C1, respectively, while features 118 and 326 are
unclustered at the correlation threshold used in Section \ref{sec:supp-cluster-scenes}.

The metrics also split into relating to the generation of trajectories and/or scoring of trajectories.
Lateral displacement, curvature, braking, and proposal spread characterize
generated controls or trajectories, whereas score margin characterizes the
proposal scorer. Because the intervention is applied to the shared planner
query, this comparison measures differential output responses rather than
strictly head-isolated causal effects.

\subsection{Cluster-Level Steering Effects}
\label{sec:supp_cluster_steering}

We average each control score over the features assigned to each co-activation
cluster. The mean control scores are $0.0794$, $0.0799$, $0.0647$,
and $0.0671$ for the best feature in each cluster C0--C3, respectively. C1 has the strongest mean
proposal spread response ($0.054$), while C0 exhibits the strongest clustered
score margin response ($0.031$).

\begin{figure}[htpb]
    \centering
    \includegraphics[width=\linewidth]
    {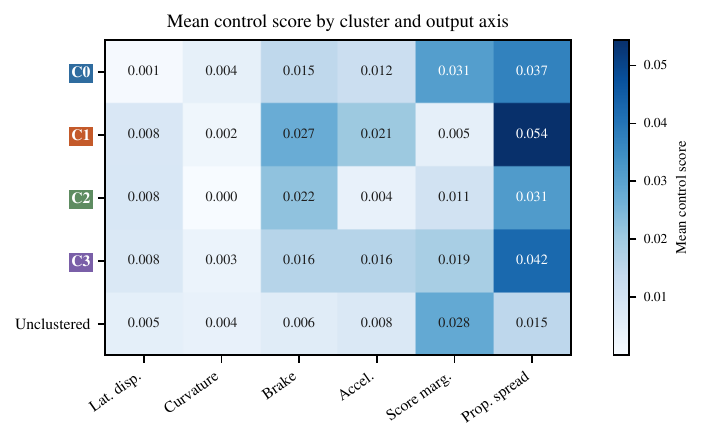}
    \caption{Mean control score for each block 3 feature cluster and planner
    statistic. Clusters differ in their secondary controlled statistics.}
    \label{fig:supp_cluster_control_heatmap}
\end{figure}

\subsection{Natural Association Versus Steering Response}
\label{sec:supp_natural_vs_steering}

We additionally compare causal steering responses with associations observed
under unedited validation inputs. For each feature and statistic, the natural
response is the Pearson correlation between feature activation and the
statistic over all validation scenes. The steering response is the
standardized effect $E_{jm}$ produced by the positive intervention. We report
results over the 270 features that activate in at least 64 validation scenes.

\begin{table}[htpb]
\centering
\small
\begin{tabular}{lcrrr}
\toprule
Statistic & Sign match & Pearson & Spearman \\
\midrule
Lateral displacement & 59.7\% & 0.686 & 0.349 \\
Curvature            & 40.3\% & $-0.289$ & $-0.025$ \\
Brake magnitude      & 56.7\% & 0.434 & 0.368 \\
Score margin         & 58.1\% & 0.233 & 0.283 \\
Proposal spread      & 61.5\% & 0.231 & 0.352 \\
\bottomrule
\end{tabular}
\vspace{2mm}
\caption{Agreement between naturally occurring activation and output associations
and positive intervention responses. Pearson and Spearman columns
measure the relationship across features between natural association strength
and the effect of intervening on SAE features.}
\label{tab:natural_vs_steering}
\end{table}

Natural associations predict intervention direction most strongly for lateral
displacement. Braking, score margin, and proposal spread also show positive,
but moderate, agreement. Curvature is the main exception, consistent with its
greater numerical sensitivity. The exhaustive sweep evaluates only positive
latent edits because the top-$k$ SAE latents are nonnegative; we therefore do
not infer symmetric negative-edit behavior from these results.

  \section{Scorer Adapter Details and Ablations}
  \label{sec:supp-scorer-adapter}
  We train a lightweight \emph{scorer adapter} that corrects the frozen
  planner's proposal-ranking scores using internal planner representations.
  The adapter does not change trajectory generation: it reranks the same
  $K{=}50$ proposals produced by the base model by adding offsets to the 
  generated scores.
  All results are evaluated on the entire WOD-E2E validation split, and 
  metrics are computed over all $50$ proposals
  per prediction. \textbf{ADE@5s} is the ADE of the scorer's top-1 pick
  (lower is better). \textbf{Mean rank} is the oracle rank (1 = best) of the
  selected proposal, averaged over scenes (lower is better). \textbf{Relative
  improvement} is $(\mathrm{ADE}_{\mathrm{base}} -
  \mathrm{ADE}_{\mathrm{method}}) / \mathrm{ADE}_{\mathrm{base}} \times 100\%$.
  \subsection{Adapter Architecture}
  \label{sec:supp-scorer-arch}
  \paragraph{Setup.}
  Let the frozen planner produce, for each scene $i$, a set of $K{=}50$
  proposal trajectories $\{\boldsymbol{\tau}_{ik}\}_{k=1}^{K}$ with
  $\boldsymbol{\tau}_{ik} \in \mathbb{R}^{2T}$, $T{=}20$ (5\,s horizon at
  $\Delta t{=}0.25$\,s), and corresponding base scorer predictions
  $\hat{s}_{ik} = S_\theta(\boldsymbol{\tau}_{ik}, q_i)$, where $q_i$ is the
  final query token. The true per-proposal ADE is
  $a_{ik} = \mathrm{ADE}(\boldsymbol{\tau}_{ik}, \mathbf{y}_i)$.
  We hook the planner at block $b{=}3$ and capture the dense
  activation $h_i \in \mathbb{R}^{d}$ with $d{=}384$. Our frozen SAE maps this activation to sparse latents using the same preprocessing and TopK encoder as in all other experiments:
\begin{equation}
  z_i =
  \mathrm{ReLU}\!\left(
    \mathrm{TopK}(W_{\mathrm{enc}}\tilde{h}_i, k)
  \right)
  \in \mathbb{R}^{M},
\end{equation}
where $\tilde{h}_i$ is the preprocessed activation obtained by
subtracting the learned preprocessing bias, removing the
per-sample mean across feature dimensions, and normalizing
to unit $L_2$ norm, as described in the implementation details.
  \paragraph{Adapter.}
  For each proposal $(i,k)$, the adapter $f_\phi$ takes three inputs:
  \begin{enumerate}
    \item the SAE latent $z_i$ (scene-level internal representation),
    \item the flattened proposal $\boldsymbol{\tau}_{ik} \in \mathbb{R}^{40}$,
    \item the base score $\hat{s}_{ik} \in \mathbb{R}$.
\end{enumerate}
  We embed each input into a hidden dimension of $256$ and pass them
  through a two-layer MLP head to predict a scalar \emph{scorer residual}
  \begin{equation}
    \hat{r}_{ik} = f_\phi(z_i,\, \boldsymbol{\tau}_{ik},\, \hat{s}_{ik}).
  \end{equation}
  The corrected score is
  \begin{equation}
    \tilde{s}_{ik} = \hat{s}_{ik} + \hat{r}_{ik},
  \end{equation}
  and the adapter selects $\arg\min_k \tilde{s}_{ik}$. The planner backbone
  and SAE are frozen throughout; only $\phi$ is trained.
  \paragraph{Training objective.}
  The regression target is the true scorer error
  $r_{ik} = a_{ik} - \hat{s}_{ik}$.
  We minimize an $L_1$ loss over proposals:
  \begin{equation}
    \mathcal{L}_{\mathrm{reg}}
    = \frac{1}{BK}\sum_{i,k} w_{ik}\,\bigl|\hat{r}_{ik} - r_{ik}\bigr|,
  \end{equation}
  where $w_{ik}$ is 3.0 to the top $10$ proposals ranked by ground truth ADE, and is 1.0 for the other 40 proposals.
  \paragraph{Hyperparameters.}
  We train the adapter for 8 epochs using AdamW with learning rate $10^{-3}$,
weight decay $10^{-3}$, batch size 16, and dropout $0.2$, with early stopping
on validation loss.
  \subsection{Dense Query Control}
  \label{sec:supp-scorer-dense}
  To test whether sparsity is necessary, we train an ablation that replaces
  the SAE latent $z_i$ with the raw dense hooked activation $h_i \in
  \mathbb{R}^{384}$ from the same block-3 MLP output. All other components
  are identical: same three-tower fusion head, same trajectory and base score
  inputs, same $\ell_1$ residual target, same frozen backbone, and same
  training schedule. The only difference is the input: replacing the SAE latent
  with the original query.
  This control isolates the value of SAE disentanglement: if sparse latents
  merely provide extra capacity, the dense adapter should match SAE-latent
  performance; if SAE features expose task-relevant structure, the
  SAE-latent adapter should outperform the dense control.
  \subsection{Matched Trainable Control}
  To further test whether the SAE adapter benefits simply from the additional encoder
parameters, we replace the pretrained frozen SAE encoder with a randomly
initialized encoder of identical architecture and parameter count. We train
this encoder jointly with the adapter head on the same $L_1$ scorer-residual
objective while keeping the base planner frozen. Thus, the two variants differ
only in whether the encoder representation is learned through SAE
pretraining or directly from the downstream score-correction objective.
This randomly initialized control reduces ADE@5s by 2.86\%, compared with 3.44\% for
the frozen SAE encoder.
  \subsection{Results}
  \label{sec:supp-scorer-ablation}
  Table~\ref{tab:scorer-adapter-ablation} summarizes the ablation. The
  SAE-latent adapter improves mean rank from $16.7$ to $15.6$ and reduces
  ADE@5s by $3.44\%$ relative to the base scorer. The dense-query control
  improves over the base scorer but falls short of the SAE-latent variant,
  indicating that sparse SAE features carry information useful for scoring
  beyond what is available in the raw dense activation. The randomly initialized adapter matching the SAE's architecture also does not produce as large of an improvement over the base scorer as the original SAE latent adapter.

  These results provide evidence that SAE latents expose
  internal structure (such as object- and intent-correlated features
  identified in Section~\ref{sec:supp-edits}) that can
  be utilized to improve model performance after training without
  requiring supervised fine-tuning of the entire model.
  \begin{table}[h]
\centering
\begin{tabular}{lcc}
\toprule
Method & Selected mean rank $\downarrow$ & ADE@5s reduction $\uparrow$ \\
\midrule
Base scorer                                   & 16.7 & -- \\
Dense query adapter                           & 16.2 & 2.74\% \\
Random init encoder adapter                   & 16.1   & 2.86\% \\
SAE latent adapter                            & 15.6 & \textbf{3.44\%} \\
\bottomrule
\end{tabular}
\vspace{2mm}
\caption{Scorer adapter ablations on WOD-E2E validation. ADE values report
relative reduction from the base scorer. The dense-query control replaces
SAE latents with the original block-3 activation, while the random-init
control replaces the pretrained SAE encoder with an encoder of matched
architecture and parameter count trained on the downstream objective.}
\label{tab:scorer-adapter-ablation}
\end{table}

\end{document}